# Prediction of Final Phosphorus Content of Steel in a Scrap-Based Electric Arc Furnace Using Artificial Neural Networks

Riadh Azzaz[1], Valentin Hurel[2], Patrice Ménard[2], Mohammad Jahazi[1], Samira Ebrahimi Kahou[3], and Elmira Moosavi-Khoonsari[1*]

[1]Department of Mechanical Engineering, École de Technologie Supérieure (ÉTS)
1100 Notre-Dame Street West, Montréal, QC, Canada, H3C 1K3

[2]Finkl Steel-Sorel
100 Rue McCarthy, Saint-Joseph-de-Sorel, QC, Canada, J3R 3M8

[3]Schulich School of Engineering, University of Calgary,
622 Collegiate Pl NW, Calgary, AB, Canada, T2N 4V8
Mila, CIFAR AI Chair

* Corresponding author: Elmira.moosavi@etsmtl.ca

# Abstract

The scrap-based electric arc furnace process is expected to capture a significant share of the steel market in the future due to its potential for reducing environmental impacts through steel recycling. However, managing impurities, particularly phosphorus, remains a challenge. This study aims to develop a machine learning model to estimate the steel phosphorus content at the end of the process based on input parameters. Data were collected over two years from a steel plant, focusing on the chemical composition and weight of the scrap, the volume of oxygen injected, and process duration. After preprocessing the data, several machine learning models were evaluated, with the artificial neural network (ANN) emerging as the most effective. The best ANN model included four hidden layers. The model was trained for 500 epochs with a batch size of 50. The best model achieves a mean square error (MSE) of 0.000016, a root-mean-square error (RMSE) of 0.0049998, a coefficient of determination ($R^2$) of 99.96%, and a correlation coefficient (r) of 99.98%. Notably, the model achieved a 100% hit rate for predicting phosphorus content within ±0.001 wt% (±10 ppm). These results demonstrate that the optimized ANN model offers accurate predictions for the steel final phosphorus content.

# 1. Introduction

Steelmaking is currently a major contributor to $CO_2$ emissions, but it is committed to advancing a sustainable metallurgical industry, as reflected in its adoption of scrap-based electric arc furnaces (EAF). This process involves melting scrap steel by generating an electric arc between electrodes and the liquid steel bath [1]. It effectively recycles steel scrap, reducing $CO_2$ emissions by 90% and energy consumption by 70% compared to the traditional blast furnace-basic oxygen furnace (BF-BOF) route. Additionally, it significantly lowers the consumption of natural resources like iron ore, coal, and limestone [2, 3].

Despite the environmental benefits of EAFs, they face complex scientific and technical challenges, particularly in managing impurities such as phosphorus (P) in steel [4, 5]. An uncontrolled quantity of P in steel negatively impacts the mechanical properties of steel, leading to increased temper and intergranular embrittlement and cracking [6, 7]. To meet quality standards, it is crucial to reduce P levels from typically above 0.025 wt% in scrap to less than 0.015 wt% in the final product. For certain applications, the target phosphorus content may need to be as low as 0.005 wt% [8]. The varied composition of scrap feedstock in comparison to ore-based production further complicates this reduction process, and steelmaking needs to align its operation continuously with the complicated composition of modern steel products [9-11].

Numerous studies have investigated P removal from steel, focusing on P equilibrium distribution and phosphate capacity at laboratory or intermediate scales [12-26]. Additionally, plant trials in EAFs have examined P behavior during direct reduced iron (DRI) and hot briquetted iron (HBI) processes [27-29]. While experimental methods are valuable, they are often time-consuming, costly, and difficult to apply on an industrial scale. Furthermore, P measurements in controlled lab conditions do not easily translate to large-scale environments where fluid flow and kinetic conditions differ significantly. Consequently, modeling and simulation provide viable alternatives to purely experimental methods, primarily divided into phenomenological (mechanistic) models based on physical phenomena, such as computational fluid dynamics (CFD) [30, 31], and data-driven statistical models [32].

Mechanistic models have greatly enhanced our understanding of the EAF process, but they come with significant limitations. For example, CFD models include reliance on equilibrium models for metal-slag-gas interactions, which sacrifice accuracy for speed, and statistical turbulence modeling that may introduce errors in unsteady flow conditions. Additionally, the use of empirical constants for mass transfer coefficients limits generalizability, while inadequate validation of foamy slag models and the oversimplification of local conditions reduce overall accuracy. The assumption of arc plasma as a black body for heat transfer may be also an oversimplification, and the computational demands of comprehensive models make them impractical for online applications [30, 32]. Consequently, while phenomenological models show promise, they still face substantial challenges, particularly in capturing the wide range of scales and phenomena in such a complex EAF process.

In light of the limitations of traditional physical models, researchers have increasingly turned to statistical and data-driven approaches for predicting the final P content in steel [33-48]. The machine learning (ML) methods offer a faster, cheaper, and safer alternative to plant trials [32] and adapt well to variations in scrap composition and operating conditions, often outperforming mechanistic models in accuracy [32, 46].

While only a few studies have focused on predicting and optimizing the EAF process [34, 45, 47, 49, 50], particularly regarding endpoint P content [34, 44, 47], most research has concentrated on the BOF process [33, 35-39, 41-43, 45, 46, 51]. As EAFs gain prominence and scrap recycling becomes crucial, accurate P prediction is vital. Existing ML models show promise, but their effectiveness varies due to data availability and quality, input parameter selection, and model robustness [45, 52]. Notably, Yuan et al. [34] developed a least squares support vector machine (LS-SVM) model that achieved an 87% hit rate for predicting P levels with ±0.003 wt% errors in EAF steel. Chen et al. [44] developed a back propagation neural network-decision tree (BPNN-DT) model with 6 hidden layers, 18 input parameters, and 50 neurons to predict the final P content of steel in EAF. The proposed hybrid model combines k-means clustering, BPNN, and a DT algorithm for prediction. The model achieves a phosphorous prediction accuracy of 97.8% within ±0.006 wt% error, 94.2% and 83.0% for ±0.005 wt% and ±0.004 wt% error ranges, respectively. Zou et al. [47] used an BPNN model, attaining a hit rate of 87.8% for ±0.004 wt% errors and 75.6% for ±0.003 wt% errors.

This study aims to develop an ML model to accurately predict the ultimate phosphorus content of steel based on key input parameters from a scrap-based EAF process. ANNs are particularly effective at modeling complex, non-linear relationships in data, which is essential for capturing the intricacies of EAF processes and accurately predicting final P content. The approach involves preprocessing original production data from a steelmaking plant to remove outliers and conducting a correlation analysis between input parameters and the target output variable. Moreover, the ANN model is compared alongside other models developed in this work, including random forest (RF) and SVM with a radial basis function (RBF) kernel, as well as models reported in the literature. The comparison employed various evaluation metrics to gauge the performance of each model.

## 2. Analysis of scrap-based EAF

### 2.1. Description of EAF process

An EAF operates in batch tap-to-tap cycles, consisting of the following steps: initial charging (3 minutes), primary melting (20 minutes), additional charging (3 minutes), secondary melting (14 minutes), refining (10 minutes), deslagging and tapping (3 minutes), and furnace tilting (7 minutes). Modern operations aim to complete the entire tap-to-tap cycle in under 60 minutes [53]. A schematic of EAF steelmaking is shown in Figure 1.

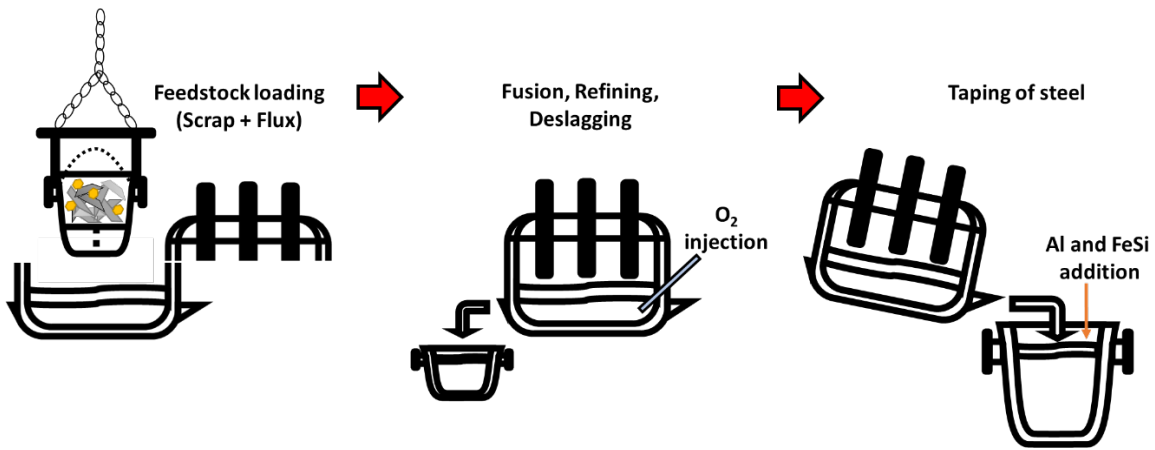

Figure 1. Schematic of an electric arc furnace steelmaking process.

**Charging the Furnace.** The furnace is charged from the top and many companies combine lime and carbon addition in the scrap basket and use additional injections as needed [54]. The number of scrap buckets used is based on furnace volume and scrap density, with modern designs aiming to minimize recharging to reduce downtime and energy loss. Typically, companies aim for 2 to 3 scrap buckets per cycle [53].

**Melting Scrap.** Melting scrap in an EAF primarily relies on electrical energy supplied by graphite electrodes. Initially, an intermediate voltage is used until the electrodes penetrate the scrap, after which a higher voltage stabilizes the arc for efficient heat transfer and forms a liquid metal pool. Chemical energy, provided by oxy-fuel burners and oxygen lances, further aids the melting process through flame radiation, convection, and exothermic reactions. The process continues with repeated charging until all scrap is melted [55].

**Refining.** Once the bath temperature stabilizes, chemical analysis directs refining operations such as oxygen blowing and alloy additions. Oxygen injection begins before stabilization, initiating some reactions early. Adjustments are made to manage excess elements like phosphorous, carbon, silicon, and chromium by transferring them to the slag phase. However, the EAF's impurity removal capacity is limited due to the lower basicity and mass of the slag. Initial slagging is crucial for removing phosphorous before reversion occurs. The final bath composition is carefully managed to meet steel specifications, with alloy additions made in the ladle to adjust the composition as needed [42, 53, 55, 56].

**Deslagging.** The slag collecting the undesired species like phosphorous is removed during the deslagging step by tilting the furnace backward and allowing the slag to exit through a designated door. This removal process reduces the risk of phosphorous reversion when the temperature is increased for further refining, such as during desulfurization or carbon injection, and during slag foaming to reduce iron oxide to metallic iron [53, 57].

**Tapping.** Tapping molten metal from a furnace is a crucial operation, and any failure in this process necessitates a complete shutdown. Operations can only resume once tapping is successfully completed. Key factors to manage during tapping are the rate and duration. It is also important to note that the furnace is never entirely emptied; a small amount of molten metal remains inside when the tapping hole is sealed [58].

## 2.2. Phosphorous removal

The phosphorous removal process can be divided into two main stages [59]. Initially, P in iron-based melts is oxidized by $Fe_tO$, which is primarily generated from the reaction of scrap with injected oxygen, forming $P_2O_5$ according to the following reaction, Equation (1):

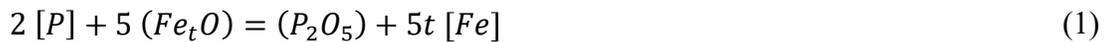

$$2\,[P] + 5\,(Fe_tO) = (P_2O_5) + 5t\,[Fe] \tag{1}$$

where [ ] and ( ) denote the species in the metal and slag phases, respectively.

Next, the injected flux (CaO) stabilizes the extracted phosphorus ($P_2O_5$) in the slag, resulting in the formation of calcium-phosphate ($3CaO \cdot P_2O_5$) through the following reaction, Equation (2):

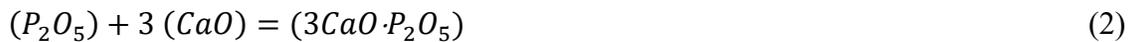

$$(P_2O_5) + 3\,(CaO) = (3CaO \cdot P_2O_5) \tag{2}$$

The phosphorus removal reaction can also be represented in its ionic form, Equation (3) [25]:

$$[P] + \frac{5}{2}[O] + \frac{3}{2}(O^{2-}) = (PO_4^{3-}) \quad (3)$$

where [P] and [O] represent phosphorus and oxygen, respectively, and $O^{2-}$ and $PO_4^{3-}$ represent the oxide and phosphate ions, respectively.

Two concepts, phosphorus partition coefficient ($L_p$) and phosphate capacity ($C_{PO_4^{3-}}$), have been developed to quantify the phosphorus removal process [25]. The $L_p$ parameter can be described as follows:

$$L_p = \frac{(\%\,P)}{[\%\,P]} \quad (4)$$

where (%P) and [%P] represent the phosphorus concentrations in the slag and steel, respectively. The $L_p$ parameter ranges from 5.0 to 15.0. Generally, phosphorus content is only reduced by about 20 to 50% during EAF treatment. However, given the low phosphorus content of scrap compared to hot metal (produced from iron ore treatment in the BF), this degree of removal is considered satisfactory [53].

The $L_p$ parameter between the slag and liquid steel is commonly used to evaluate the phosphorus removal capability of the slag due to its ease of measurement in both laboratory studies and commercial production [12-26, 59, 60]. Nevertheless, it is crucial to note that this ratio can only be used as a comparative measure between different slag compositions if the partial oxygen pressure ($P_{O_2}$) is equivalent in the compared systems [61].

Wagner proposed the concept of phosphate capacity ($C_{PO_4^{3-}}$) to describe the slag's phosphorus removal potential using a slag-gas equilibrium reaction. $C_{PO_4^{3-}}$ incorporates the influence of $P_{O_2}$, making it an essential measure for the comparative evaluation of various slag systems. The slag-gas reaction and $C_{PO_4^{3-}}$ are represented by Equations (5) and (6), respectively [61]:

$$\frac{1}{2}P_2(g) + \frac{5}{4}O_2(g) + \frac{3}{2}O^{2-} \text{ (slag)} = PO_4^{3-} \text{ (slag)} \tag{5}$$

$$C_{PO_4^{3-}} = \frac{(\%PO_4^{3-})}{P_{P_2}^{1/2} P_{O_2}^{5/4}} = \frac{K_{(2)}(a_{O^{2-}})^{3/2}}{\gamma^{\circ}_{PO_4^{-3}}} \tag{6}$$

where %PO$_4^{3-}$ is the weight percentage of PO$_4^{3-}$ dissolved in the slag, and $P_{O_2}$ and $P_{P_2}$ are the partial pressures of oxygen and phosphorus, respectively, at the slag-gas interface in equilibrium. In cases where the concentration of PO$_4^{3-}$ is notably low, it is acceptable to replace the activity with the corresponding concentrations of (%PO$_4^{3-}$) multiplied by a critical constant parameter $\gamma^{\circ}_{PO_4^{-3}}$, representing the activity coefficient at infinite dilution. $K_{(2)}$ is the equilibrium constant for reaction (5).

The $C_{PO_4^{3-}}$ shows a direct correlation with $L_p$ as shown in Equation (7):

$$C_{PO_4^{3-}} = \frac{L_p k_p}{f_p P_{O_2}^{5/4}} \tag{7}$$

where $k_p$ is the equilibrium constant for phosphorus dissolution in iron ($\frac{1}{2}P_{2(g)} = [P]$) [61].

## 2.3. Factors influencing phosphorus removal

Several factors influence the effectiveness of phosphorus removal in the EAF. Key parameters affecting phosphorus elimination include the slag's basicity, temperature, and FeO content. From a thermodynamic point of view, low temperatures, high FeO content, and increased basicity generally favor the phosphorus removal process [19-22, 62, 63].

**Basicity.** The basicity of slag is usually expressed as the molar ratio of basic oxides (e.g., CaO) to acidic oxides (e.g., $SiO_2$). It is a critical factor in metallurgy, influencing the slag's ability to absorb impurities like phosphorus, as well as its melting point and viscosity [64, 65]. Increasing the basicity of the slag, typically by raising the concentration of basic oxides such as CaO, enhances phosphorus removal efficiency by stabilizing $P_2O_5$ as $4CaO·P_2O_5$ at steelmaking temperatures. However, overly high basicity can be counterproductive. Excessive basicity raises the slag's melting point, preventing complete melting of CaO particles and increasing slag viscosity. This increased viscosity reduces the phosphorus diffusion in the slag, slowing the phosphorus removal reaction at the interface between the molten steel and slag and thus diminishing removal efficiency [25, 38, 62].

**FeO Content**. The effectiveness of phosphorus removal in CaO-based slags is also influenced by the presence of iron(II) oxide (FeO), which can act as an acidic or basic oxide depending on the slag composition and oxygen potential. Research has shown that the phosphorus removal capacity, or phosphate capacity, of $CaO-SiO_2-MgO-FeO$ slags increases with FeO content. Lee and Fruehan [20] observed an increase in the phosphate capacity with FeO content between 3 and 10 wt% at high temperatures. Hamano and Tsukihashi [19] found a maximum phosphate capacity at about 50 wt% FeO which then decreases with further increasing the FeO content to 60 wt%. Li et al. [21] noted that the phosphate capacity peaks at 25-35 wt% FeO and then decreases, which is attributed to the dilution of CaO, reducing its activity and increasing the activity of $P_2O_5$ [62]. Thus, optimizing FeO content is crucial for effective phosphorus removal in CaO-based slags.

**Temperature.** Temperature impacts phosphorus removal in two contrasting ways. High temperatures can negatively affect the process because phosphorus removal is highly exothermic. Conversely, elevated temperatures promote the melting of lime, which enhances the basicity of the slag. This improved basicity aids in the distribution of phosphorus into the slag phase and increases the $L_p$, thereby enhancing removal efficiency. On the other hand, temperature favors the kinetics of the phosphorus removal process [38, 48].

## 3. Prediction of End-Point Phosphorus Content in Steel

### 3.1. Machine learning algorithms

A wide range of ML models has been employed in steel dephosphorization for process prediction and optimization. Zou et al. [47] investigated the EAF process combining scrap and hot metal (HM) with a BPNN using 10 parameters on a dataset of 1,250 (580 utilized). Chen et al. [44] implemented a k-means-BPNN-DT model with 18 parameters, analyzing a dataset of 1,258 (1,114 utilized). Yuan et al. [34] applied least squares support vector machine with principal component regression (LS-SVM-PCR) with 10 parameters on a smaller dataset of 82. Zhang et al. [46] focused on the BOF process, utilizing multiple models including Ridge Regression, gradient boosting regression (GBR), SVM, RF, and convolutional neural network (CNN) on a substantial dataset of 13,000 (7,776 utilized).

Zhou et al. [51] examined both unconstrained and monotone-constrained BPNN models, utilizing a dataset of 900. Wang et al. [48] evaluated an unhybrid ANN and a hybrid physics-based ANN with 19 parameters on a dataset of 28,000. Chang et al. [43] assessed various models including partial least squares (PLS), SVR, fully convolutional network (FCN), extreme learning machine (ELM), graph convolutional network (GCN), and multi-channel GCN. They used 42 parameters. He and Zhang [41] employed principal component analysis (PCA) and BPNN on a dataset of 1,978. Liu et al. [39] utilized computer vision and General Regression Neural Networks (GRNN) in a BOF process. Wang et al. [66] applied multi-level recursive regression in a BOF context, leveraging 4 input variables and analyzing a dataset of 1450 samples. Additionally, Wang et al. [36] employed a BP-ANN in the BOF process, using 21 input variables with a dataset of 2000 samples.

In this study, three specific regression models are employed: RF, SVM, and ANN. In general, SVM and RF are simpler than ANN and require less training time, which is why they were tested first. However, due to the need for higher prediction accuracy, we ultimately developed an ANN model, as the accuracy of the former models was insufficient. A detailed presentation of these techniques will be provided in the following section.

### 3.1.1. Random Forest

Random Forest can function as both a regressor and a classifier. Its primary role is to enhance prediction accuracy and prevent overfitting by constructing multiple decision trees from different subsets of the dataset and averaging their predictions. Key hyperparameters that influence the RF model include the number of trees, the criteria for splitting nodes (e.g., minimum node size), and the degree of randomness introduced. Although tuning these hyperparameters can optimize performance, the RF model generally performs well with default settings provided in software packages [67]. For this study, the RF model was implemented using the scikit-learn library, facilitating easy model execution and result analysis.

### 3.1.2. Support Vector Machine

Support vector machine is renowned for its strong generalization capabilities and is extensively utilized for regression tasks due to its high prediction accuracy. The SVM training process leverages Vapnik's principle of structural risk minimization to mitigate generalization error [68]. Key hyperparameters for the SVM algorithm include Kernel, Gamma ($\gamma$), and regularization parameter (C). The Kernel parameter determines the type of kernel function used, with the Radial Basis Function (RBF) being the default choice, effective for non-linear relationships [69]. Gamma controls the influence of a single training example, where a very high value indicates a smaller region of influence and a very small value does not allow for capturing the complexity of dataset [70]. The parameter C is the penalty factor for error terms, which governs the balance between achieving a low error on the training data and minimizing model complexity. A higher C value places greater emphasis on correctly classifying outliers. For this study, the SVM-RBF model was implemented.

### 3.1.3. Artificial Neural Network

Artificial Neural Networks are a robust ML framework known for their ability to model complex non-linear relationships which is the case in metallurgical processes. Laha et al.

[40] and Wang et al. [48] applied this model to predict the yield output of steel and steel composition, respectively. They achieve this by processing inputs through intricate interconnections between neurons distributed across multiple layers, transforming them into outputs. Each neuron functions as a fundamental unit, handling and passing information within the network. Figure 2 illustrates a basic neuron and the overall architecture of an ANN model.

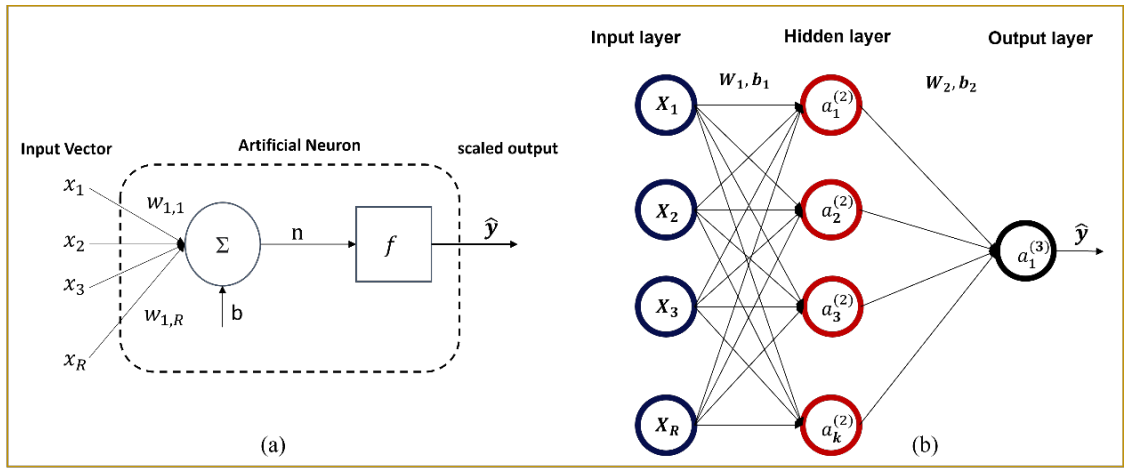

Figure 2. Architecture of an artificial neural network

A basic neural network consists of three main components: an input layer that receives data, an output layer that makes predictions, and one or more hidden layers that process information through interconnected neurons. Each neuron in the hidden layers operates with weights and biases, as described by Equation (8):

$$a = f\left[\sum_{i=1}^{n} w_i + b\right] \tag{8}$$

where $w_i$ and b represent the weights and bias values, respectively, while $x_i$ denotes the inputs and $f[.]$ denotes the activation function. During training, the network learns and adjusts these weights to optimize performance. The use of activation functions enables ANNs to learn complex patterns, making them capable of universal approximation—mapping any input to any output regardless of data complexity [71].

*3.1.3.1. Establishment of artificial neural network models*

Figure 3 illustrates the sequence of steps involved in developing an ANN model for this study. The process begins with data collection and preprocessing, including tasks such as data cleaning, correlation analysis, and normalization. Next, the data is split into three distinct sets: a training set, a validation set, and a test set. The final stage involves training the model, where an appropriate neural network architecture is selected, and key hyperparameters—such as the number of layers, the number of neurons per layer, and the activation function—are identified and fine-tuned.

Three categories of data sets are used: training, validation, and test sets. The training set provides information on the target function to train the network. The validation set is used in conjunction with early stopping techniques to monitor and prevent overfitting by tracking validation errors during training. After training, the test set is employed to evaluate the model's performance. Typically, 60% of the data is allocated for training, while 40% is divided between validation and testing. Of this, 20% is reserved for an independent generalization test of the ANN. To ensure unbiased evaluation, the validation set is drawn from the training data after shuffling, ensuring it is representative of the overall dataset.

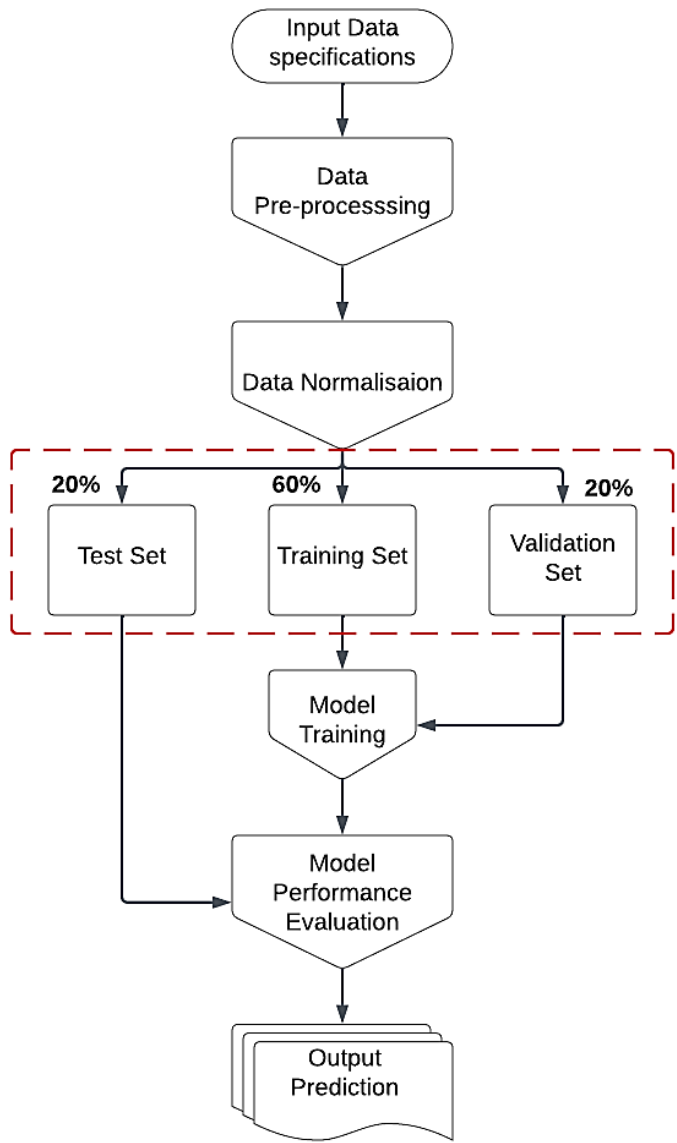

Figure 3. Flow chart for the development of the artificial neural network model based on historical plant data.

The number of hidden layers and nodes within these layers is crucial for determining the performance of an ANN model. Generally, the complexity of the network's structure is influenced by the number of hidden layers. In this study, three different ANN architectures were tested, each varying in the number of hidden layers and nodes. All architectures utilized the sigmoid activation function, which generates an output value based on the input value of the neuron, as shown in Equation (9). These models were implemented using the TensorFlow library. The mean squared error (MSE) was used as the loss function for training, as detailed in Equation (10).

$$f(y) = \frac{1}{1 + e^{-y}} \tag{9}$$

$$\text{Loss}(y, \hat{y}) = \frac{1}{N} \sum_{j=1}^{N} (y_j - \hat{y}_j)^2 \tag{10}$$

where $\hat{y}$ represents the predicted value, and $y$ denotes the actual value.

To optimize hyperparameters in an ANN model, three approaches have been proposed: grid search, random search, and manual trial-and-error [72, 73]. In this study, hyperparameters were optimized using a trial-and-error approach, testing a range of values to identify the optimal settings. The selection was based on achieving the lowest validation loss, measured by MSE. While each technique has its advantages and disadvantages, trial-and-error can be an effective alternative to adaptive (sequential) hyperparameter optimization algorithms [73]. It offers quick feedback, and is straightforward to implement, requiring no prior knowledge of complex optimization methods.

During training, the Adam optimizer was employed to adjust the learning rate adaptively for each model parameter, contrasting with traditional stochastic gradient descent by managing individual learning rates. This approach facilitates faster convergence and better adaptation to variations in the training data.

### 3.2. Data treatment

The present study focuses on a 40-ton EAF equipped with three graphite electrodes and charged with two scrap bins. Initially, scrap from the first bin is melted in the superheated furnace, followed by the addition of scrap from the second bin. Chemical analysis of the steel is conducted at two critical stages: before deslagging and just before transferring the liquid metal to the ladle furnace (LF) at 1650°C. The second analysis is crucial because, with the slag removed, phosphorus may revert into the steel, making final phosphorus content a key parameter for process control.

The steelmaking process produces a large volume of data, but this raw data often contains missing values, outliers, and inconsistencies, which can significantly impact model performance if used directly. Thus, preprocessing is essential to refine and prepare the data for ML. The methodology for data preprocessing varies based on the quality and nature of the raw data. The following section will provide a comprehensive overview of the data preprocessing stages used in the present study.

### 3.2.1. Data collection

In this study, over 1,700 heat data sets were collected from a steel plant over a two-year period. These data sets include a range of variables, such as the chemical composition of scrap and various process parameters. Table 1 outlines the parameters used to develop the ANN models, including their symbols and the rationale for their selection. Twelve parameters were chosen based on principles of metallurgy, thermodynamics, and current industrial practices [19-22, 62, 63, 74-78]. These parameters include the weight and composition of the scrap (C, Mn, Cr, Si, and S), the quantities of injected oxygen and lime, energy consumption, deslagging and tapping temperatures, and process duration.

Table 1. Input parameters used to develop the machine learning models

| Variables | Description of variables | Justification |
|---|---|---|
| $x_1$ | Scrap weight | Main material of EAF (Source of P) |
| $x_2$ | C content in scrap | |
| $x_3$ | Mn content in scrap | Elements in scrap affecting dephosphorization |
| $x_4$ | Cr content in scrap | |
| $x_5$ | Si content in scrap | |
| $x_6$ | S content in scrap | |
| $x_7$ | Injected oxygen | Oxidant |
| $x_8$ | Injected lime | Dephosphorization agent |
| $x_9$ | Energy consumption | |
| $x_{10}$ | Deslagging temperature | Process Parameters |
| $x_{11}$ | Tapping temperature | |
| $x_{12}$ | Process duration | |

### 3.2.2. Data cleaning

In this work, data cleaning was performed after data collection to address issues such as missing or aberrant values. Understanding the distribution of the data, including central tendency, dispersion, and potential outliers, is crucial for making informed decisions. Box plots are used to graphically represent these characteristics in the case of processes where strong variabilities are observed, and the raw data were analyzed using the boxplot concept, as shown in Figure 4. This method, as described by Dovoedo and Chakraborti [79], is employed to identify outliers. This method relies on five key statistics: the first quartile (Q1), the median (Q2), the third quartile (Q3), and the interquartile range (IQR), which is the difference between Q3 and Q1. Outliers are defined as values falling below Q1 - 1.5 IQR or above Q3 + 1.5 IQR. In this study, outliers were removed based on these criteria, and the cleaned data distribution is illustrated after normalization in Figure 5.

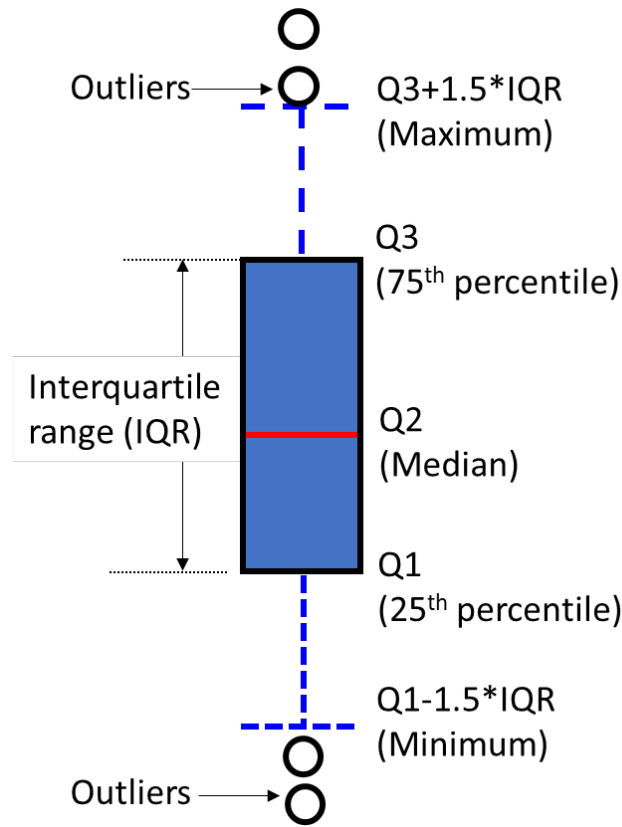

Figure 4. Key features of a box plot diagram for identifying outliers and understanding data distribution.

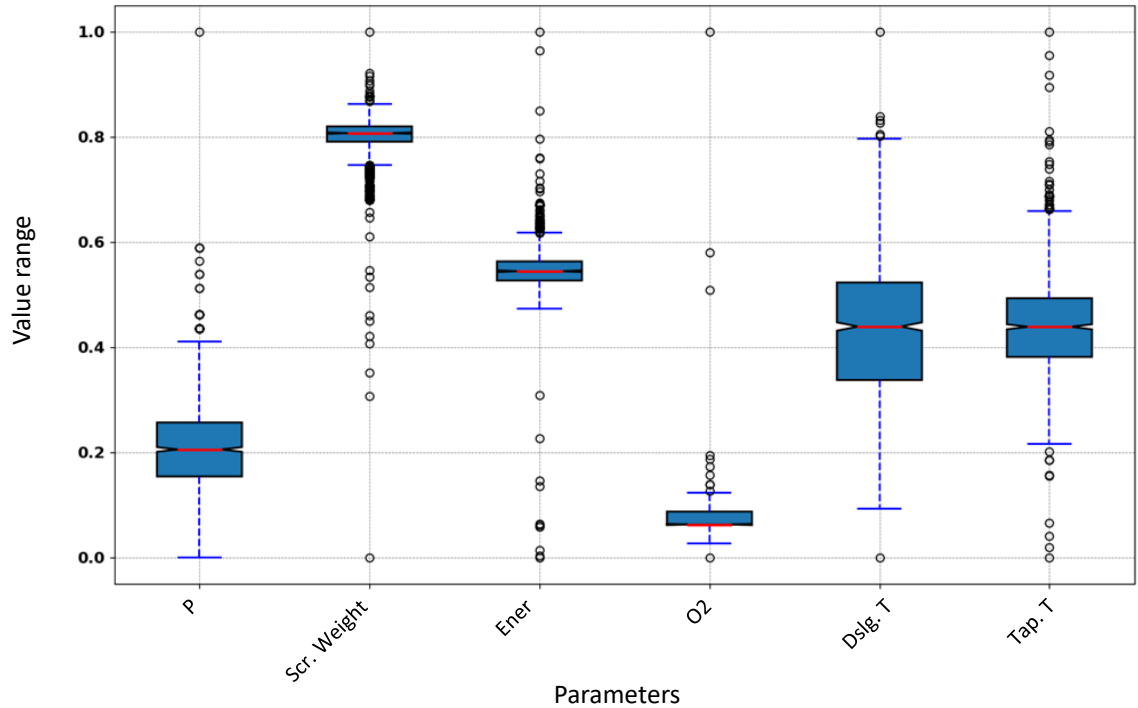

Figure 5. Descriptive statistics for the various parameters, including minimum and maximum values, mean, standard deviation, and identification of outliers for each parameter. Durat.: Process duration; Scr. Weight: Scrap weight; Tap. T: Tapping temperature; Dslg. T: Deslagging temperature; Ener: Energy consumption.

Following preprocessing and outlier elimination, approximately 1,005 data were retained. The descriptive statistics for all input and output variables of the prediction models are presented in Table 2. The range of final phosphorus content in steel varied from 0.003 to 0.018 wt% (~0.0098 wt% ± 0.0028). The carbon content in scrap ranged from 0.058 to 0.345 wt% (~0.2747 wt% ±0.0489), while the manganese content varied from 0.577 to 3.58 wt% (~0.7980 wt% ±0.0992). The chromium content in scrap ranged from 0.112 to

1.878 wt% (~0.7456 wt% ±0.2646), and the silicon content varied between 0.128 and 0.79 wt% (~0.2345 wt% ±0.0362). Sulfur content in scrap ranged from 0.004 to 0.08 wt% (~0.0127 wt% ±0.0033).

For process parameters, injected oxygen levels ranged from 77.87175 to 289.96608 m$^3$ (2,750 to 10,240 ft$^3$), and injected lime varied from 975.224 to 1,950.447 kg (2,150 to 4,300 lb). Energy consumption ranged between 18,008 and 23,398 kWh. The deslagging temperature ranged from 1,518 to 1,682°C, while the tapping temperature varied from 1,609 to 1,696°C. Scrap weight ranged from 41,340 to 43,708 kg.

Table 2 Statistics describing the input and output variables for prediction.

| Feature category | Feature | Min value | Max value | Mean | STD* |
|---|---|---|---|---|---|
| Endpoint | P content in steel (wt%) | 0.003 | 0.018 | 0.0098 | 0.0028 |
| Scrap key composition | C content in scrap (wt%) | 0.058 | 0.345 | 0.2747 | 0.0489 |
| | Mn content in scrap (wt%) | 0.577 | 3.58 | 0.7980 | 0.0992 |
| | Cr content in scrap (wt%) | 0.112 | 1.878 | 0.7456 | 0.2646 |
| | Si content in scrap (wt%) | 0.128 | 0.79 | 0.2345 | 0.0362 |
| | S content in scrap (wt%) | 0.004 | 0.08 | 0.0127 | 0.0033 |
| Process parameters | Injected oxygen (m$^3$) | 77.87175 | 289.96608 | 179.04839 | 29.95939 |

| | | | | |
|---|---|---|---|---|
| Injected lime (kg) | 975.224 | 1,950.447 | 1,047.79838 | 256.279689 |
| Energy consumption (kWh) | 18008 | 23398 | 20702 | 941 |
| Deslagging temperature (°C) | 1518 | 1682 | 1600 | 55 |
| Tapping temperature (°C) | 1609 | 1696 | 1652 | 27 |
| Scrap weight (kg) | 41340 | 43708 | 42673 | 783 |
| Process duration (min) | 98 | 1355 | 147 | 53 |

*STD: Standard deviation

### 3.2.3. Correlation analysis and normalization

*3.2.3.1 Correlation analysis*

Correlation analysis is used for understanding the relationships between independent variables and the target variable, such as the final phosphorus content in steel. This analysis clarifies the strength of associations between features and phosphorus outcomes, which is particularly valuable for ML models with simpler structures, such as RF and SVM. It has been reported that these models benefit from correlation insights to assess feature importance and guide feature selection. Feature selection is proposed for various reasons such as, improving model interpretability, reducing training time, and enhancing learning accuracy [80]. However, for more complex models like ANNs, Moosavi-Khoonsari et al. [50] found that correlation analysis can be somewhat redundant and may even lead to decreased accuracy. ANNs can automatically learn and capture intricate, non-linear relationships between features and the target variable through their multiple layers and neurons, reducing the need for explicit correlation data. Nevertheless, exploring correlations can still offer valuable insights into feature dynamics and relationships, enhancing our understanding of the data and model behavior.

The Pearson correlation coefficients ($r$) and $p$-values were utilized for correlation analysis in this study, as described in Equations (11) and (12). The $r$ values for the identified variables are depicted in Figure 5, which illustrates the linear relationships between these variables and the final phosphorus content in steel. This visualization highlights the impact of input variables on the final phosphorus content, ordered from strongest to weakest correlations, and includes both positive and negative relationships. Magnitude indicates the strength of the linear relationship between two variables. Values close to 1 or -1 signify a strong relationship, whereas values nearer to 0 indicate a weaker relationship. Positive $r$ values imply a positive relationship, meaning as one variable increases, the other also increases. Negative $r$ values suggest a negative relationship, where an increase in one variable corresponds to a decrease in the other.

Specifically, the analysis reveals that oxygen ($O_2$), sulfur (S) and manganese (Mn) content of scrap, injected lime (CaO), energy consumption, deslagging temperature, and the carbon (C) and silicon (Si) contents of scrap exhibit a negative correlation with phosphorus content. Among these, the negative correlation is strongest for oxygen and weakest for carbon. Increasing these variables generally leads to a reduction in phosphorus content of steel. Conversely, process duration, chromium (Cr) content of scrap, scrap weight, and tapping temperature show a positive correlation with phosphorus content, with process duration having the strongest and tapping temperature the weakest positive correlation. Increases in these variables generally result in higher final phosphorus content of steel.

$$r = \frac{\sum_{i=1}^{n}(x_i - \bar{x})(y_i - \bar{y})}{\sqrt{\sum_{i=1}^{n}(x_i - \bar{x})^2 \sum_{i=1}^{n}(y_i - \bar{y})^2}} \tag{11}$$

Let $\bar{x}$ represent the mean of the variable $x$; $\bar{y}$ represent the mean of the variable $y$; $x_i$ denote the $i$th value of variable $x$; and $y_i$ denote the $i$th value of variable $y$.

$$t = \frac{r\sqrt{n-2}}{\sqrt{1-r^2}} \tag{12}$$

Here, $r$ represents the correlation coefficient; $n$ denotes the sample size; and $n$-$2$ indicates the degree of freedom.

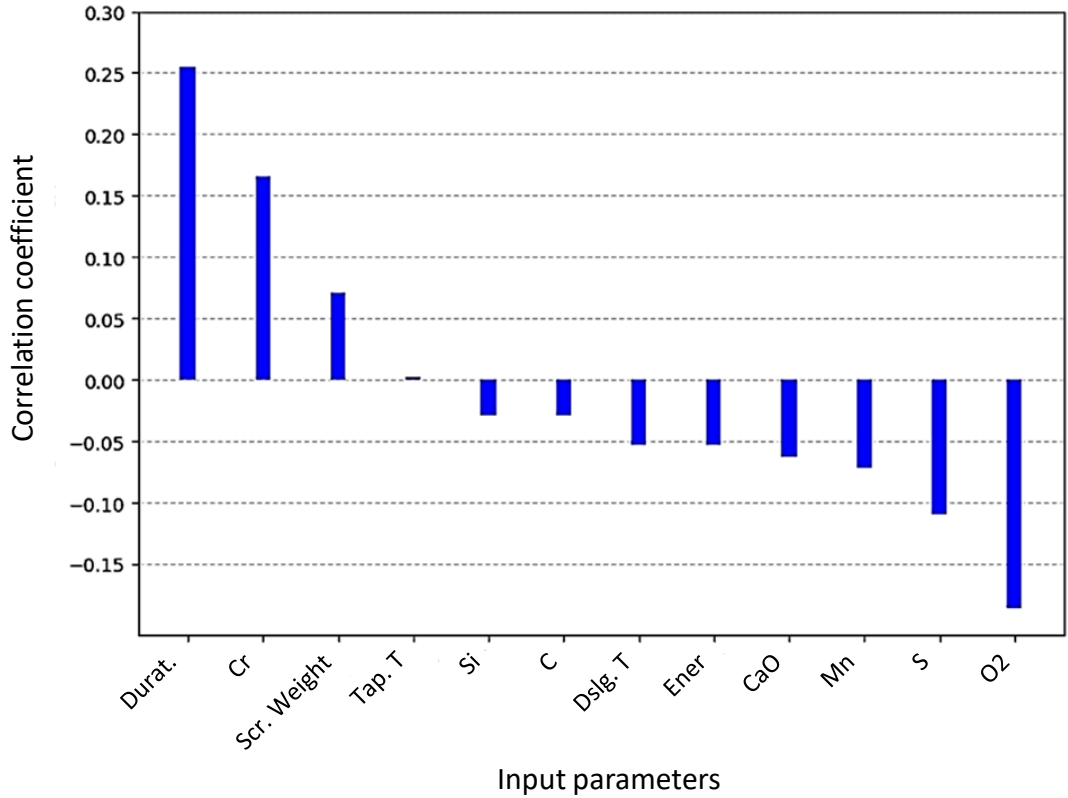

Figure 5. Pearson correlation coefficients (r) between final phosphorus content of steel (the target variable) and input parameters in electric arc furnace. Durat.: Process duration; Scr. Weight: Scrap weight; Tap. T: Tapping temperature; Dslg. T: Deslagging temperature; Ener: Energy consumption.

The p-values were analyzed to evaluate whether the correlations between the final phosphorus content in steel and the input parameters, as listed in Table 3, were statistically significant. A $p$-value less than 0.01 indicates that the correlation is very significant, suggesting a strong likelihood that the observed relationship is not due to random chance. A $p$-value less than 0.05 signifies that the correlation is significant, though less robust than those with $p$-values below 0.01. Conversely, a $p$-value greater than 0.05 implies that the correlation is not statistically significant, indicating that the relationship may be due to random variation [81]. Based on the $p$-values, the relationships between the input parameters and the final phosphorus content in steel can be categorized into three levels of significance. The most statistically significant correlations, with $p$-values less than 0.01, include the process duration ($p = 2.53\times10^{-16}$), Cr content in scrap ($p = 1.39\times10^{-7}$), injected oxygen ($p = 3\times10^{-9}$), and S content in scrap ($p = 5\times10^{-4}$). These parameters exhibit strong relationships with phosphorus content, indicating highly significant correlations. In the intermediate category, with $p$-values between 0.01 and 0.05, are scrap weight ($p = 2.44\times10^{-2}$), Mn content in scrap ($p = 2.22\times10^{-2}$), and injected lime ($p = 4.73\times10^{-2}$), suggesting these variables have notable but less pronounced effects. Finally, parameters with $p$-values greater than 0.05, including energy consumption ($p = 9.23\times10^{-2}$), deslagging temperature ($p = 9.35\times10^{-2}$), C content in scrap ($p = 3.49\times10^{-1}$), Si content in scrap ($p = 3.56\times10^{-1}$), and tapping temperature ($p = 9.38\times10^{-1}$), show weaker and statistically insignificant correlations with final phosphorus content of steel.

Table 3. Calculated results of $p$-value between final phosphorous content of steel and input variables.

| Input parameters | $r$ | $p$-value |
| --- | --- | --- |
| Scrap weight | 0.07 | $2.44 \times 10^{-2*}$ |
| C content in scrap (kg) | -0.03 | $3.49 \times 10^{-1}$ |
| Mn content in scrap (kg) | -0.07 | $2.22 \times 10^{-2*}$ |
| Cr content in scrap (kg) | 0.17 | $1.39 \times 10^{-7**}$ |
| Si content in scrap (kg) | -0.03 | $3.56 \times 10^{-1}$ |
| S content in scrap (kg) | -0.11 | $5 \times 10^{-4**}$ |

| Injected oxygen (kg) | -0.18 | $3 \times 10^{-9}$** |
| Injected lime (kg) | -0.06 | $4.73 \times 10^{-2}$* |
| Energy consumption | -0.05 | $9.23 \times 10^{-2}$ |
| Deslagging temperature | -0.05 | $9.35 \times 10^{-2}$ |
| Tapping temperature | 0.005 | $9.38 \times 10^{-1}$ |
| Process duration | 0.255 | $2.53 \times 10^{-16}$** |

1005 data were analyzed.

$p$-values < 0.05 are marked with (*), and $p$-values < 0.01 with (**).

Based on the correlation analysis, an increase in the process duration correlates with a rise in the final phosphorus content in steel. This phenomenon is attributed to the reversion of phosphorus from the slag back into the steel. Specifically, after slag removal, a thin layer of slag remains on the steel. The steel is then heated to an average temperature of 1652°C for approximately 10 minutes, though this duration is sometimes extended to 20-30 minutes [8]. Following this, the steel is transferred to the LF, conditions that are not optimal for effective dephosphorization (reducing atmosphere and high temperature). There is a direct correlation between the injected $O_2$ and the final phosphorus content in steel. The injection of $O_2$ promotes the oxidation of scrap, increasing the FeO content in the slag. As detailed in Section 2.3, the presence of FeO and CaO enhances the phosphorus removal process [19-22, 62, 63]. Increasing the amount of added CaO, decreases phosphorus content in steel by raising the slag's basicity. Conversely, an increase in chromium (Cr) leads to an increase in the final P content of the steel. Karbowniczek et al. [74] also demonstrated that increasing the chromium content in metal and the $Cr_2O_3$ content in slag results in a decrease in phosphorus removal capacity, irrespective of other parameters. Chromium in scrap oxidizes to $Cr_2O_3$ in the slag, which diminishes the dephosphorization capacity because $Cr_2O_3$ is an acidic component that reduces slag basicity. Additionally, chromium stabilizes phosphorus in the metal phase, as indicated by a negative activity coefficient ($e_P^{Cr} = -0.93$ [78], $-0.03$ [75]). Moreover, $Cr_2O_3$ promotes the formation of spinel solid particles, which reduces the proportion of liquid slag and increases its viscosity. This, in turn, lowers the dephosphorization capacity by decreasing the amount of liquid slag available for phosphorus removal and potentially slowing the kinetics of dephosphorization. Yang et al. [76] also reported that $Cr_2O_3$ levels exceeding 0.5 wt% in slag lead to increased viscosity. An increase in sulfur (S) content in scrap results in a decrease in phosphorus content in steel. This effect can be attributed to the increased activity coefficient of phosphorus in the presence of sulfur in the metal ($e_P^S = 0.028$ [75], 0.048 [78]). The oxidation of manganese (Mn) from the scrap leads to a high MnO content in the slag. MnO increases slag basicity, decreases viscosity, and lowers the liquidus temperature [77]. Consequently, it is expected to enhance the dephosphorization process, as observed in this study. An increase in scrap weight raises the phosphorus content in steel, as scrap serves as a source of phosphorus in the process.

### 3.2.3.2. Data normalization

In ML models, various parameters with distinct values and units are utilized. To facilitate the learning process and ensure rapid model convergence while mitigating bias from differing scales, it is crucial to standardize the data. The Min-Max normalization method is used for this purpose, scaling values to the range [0, 1]. The normalization process is mathematically expressed in Equation (13):

$$z_i = \frac{x_i - \min(x)}{max(x) - \min(x)} \tag{13}$$

where $z_i$ is the normalized value, $x_i$ is the original value, $max(x)$ denotes the maximum value of the data and $min(x)$ denotes the minimum value of the data.

**3.4. Model evaluation**

The efficiency of the ML models was evaluated using several statistical metrics, including MSE, RMSE, Coefficient of Determination ($R^2$), and the $r$. The mathematical formulas for calculating MSE, RMSE, and $R^2$ are provided in Equations (14-16).

$$MSE = \frac{1}{N} \sum_{j=1}^{N} (y_j - \hat{y}_j)^2 \tag{14}$$

$$RMSE = \sqrt{\frac{1}{N} \sum_{j=1}^{N} (y_j - \hat{y}_j)^2} \tag{15}$$

$$R^2 = 1 - \frac{\sum_{i=1}^{m}(\hat{y}_j - y_i)^2}{\sum_{i=1}^{m}(\bar{y} - y_i)^2} \tag{16}$$

where, $N$ is the number of the entire data set; $y_j$ is the $i$th actual value of $y$; $\hat{y}_j$ is the $i$th predicted value of $\hat{y}$; and $\bar{y}$ the average of actual values.

For regression models, the correlation coefficient $r$ can be also used to evaluate the relationship between predicted values and actual values (refer to Equation (11)). However, it is often supplemented with other metrics such as $R^2$, MSE, or RMSE for a more comprehensive evaluation.

## 4. Results and Discussion

### 4.1. Hyperparameter optimization of ANN

As mentioned in Section 3.1.3.1, the hyperparameters were optimized using the approach proposed by Begstra et al. [73, 82] to develop an optimal ANN model. The goal was to design architectures tailored to the specific problem at hand. Various combinations were tested, with different numbers of hidden layers, neurons, and iterations, while considering the required learning time. Table 4 provides a summary of the ANN structures tested. After an extensive series of trials and evaluations, a configuration that was both simple and effective was identified. The ADAM optimizer was used to manage the learning rate, facilitating faster model convergence.

Table 4. Hyperparameter used in the developed ANN models

| Hyperparameters | Different models | | |
| --- | --- | --- | --- |
| | ANN (1) 12-16-8-1 | ANN (2) 12-144-256-64-1 | ANN (3) 12-128-128-128-64-1 |
| Number of neurons | 24 | 464 | 448 |
| Number of layers | 2 | 3 | 4 |
| Number of epochs | 1000 | 100 | 500 |
| Batch size | 50 | 50 | 50 |

The first tested ANN model (ANN (1)) featured 2 hidden layers, with 16 neurons in the first layer and 8 neurons in the second layer, and was trained for 50,000 iterations, as shown in Figure 6a. The figure displays two curves: the training loss (in blue) and the validation loss (in orange), plotted against the total number of iterations. Both curves exhibit a similar pattern of gradual decline, suggesting a reduction in loss over time. After 50,000 iterations, the model achieved a MSE value of 0.0148.

A second architectural configuration (ANN (2)) was tested, featuring a more complex model with additional hidden layers and neurons compared to the previous one. This configuration included 144 neurons in the first hidden layer, 256 in the second, and 64 in the third, with a total of 5,000 iterations. As shown in Figure 6b, the number of iterations was insufficient, as the convergence curves did not stabilize. The minimum MSE achieved with this configuration was 0.0097.

The third architectural configuration (ANN (3)) was tested to improve the model's precision and overall accuracy. This iteration introduced an additional hidden layer, resulting in a total of four hidden layers, making the model moderately more complex than the previous two. In this configuration, the number of neurons per layer was reduced, with 128 neurons in the first three hidden layers and 64 in the fourth layer, and the model was trained for 25,000 iterations. As shown in Figure 6c, the MSE approached zero after approximately 6,000 iterations, with a minimum MSE of 0.00003.

As a concluding phase in the process of optimizing the ANN model, the model ANN (3), which has the following architectural parameters: 128-128-128-64, was selected. A minor alteration to the dataset was implemented by combining the validation set with the training set, thus enabling the model to be retrained with a more extensive dataset. The initial division of the total dataset into three subsets was as follows: 60% for training, 20% for validation, and 20% for testing. In the final stage of the optimization process, the dataset was divided into two subsets: 80% for training and 20% for testing. This approach ensures that the model demonstrates generalization capabilities that extend well beyond the parameters of the training data. At last, the model was evaluated based on its performance when tested with the test set. The results of the model convergence are presented in Figure 6d. It is evident that the model converges at a faster rate than previous models, reaching a minimum MSE value after just 5,000 iterations.

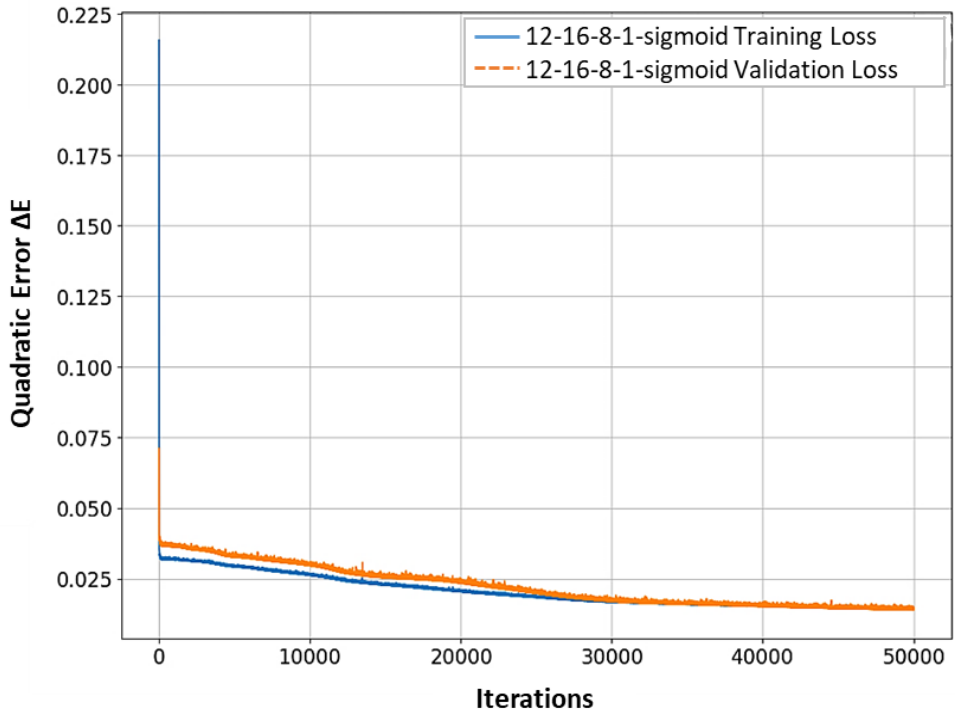

(a)

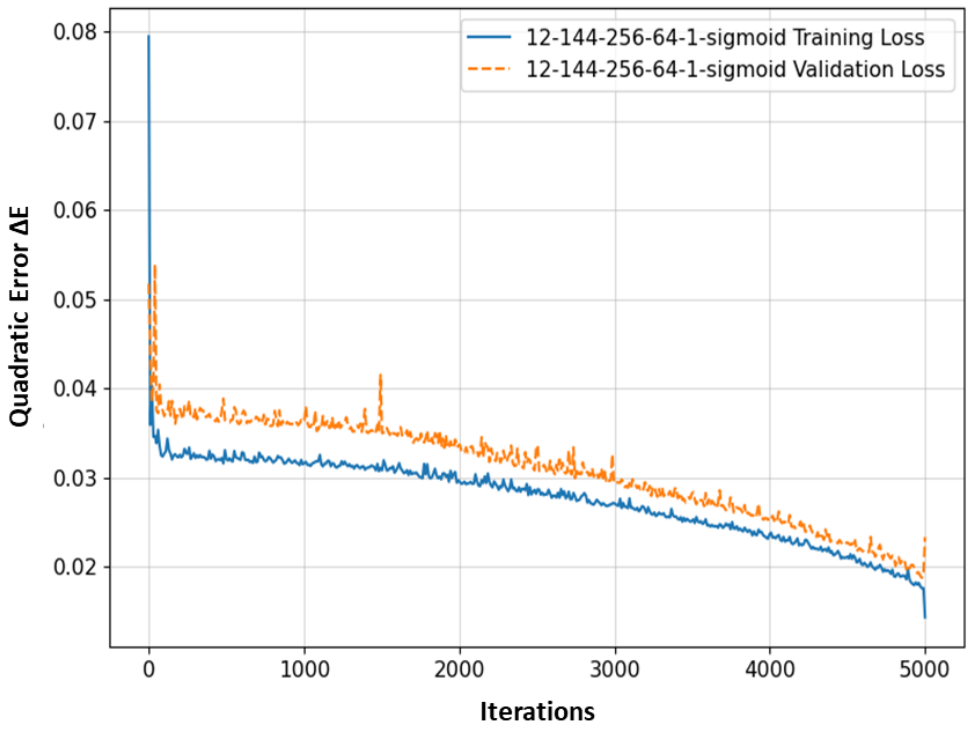

(b)

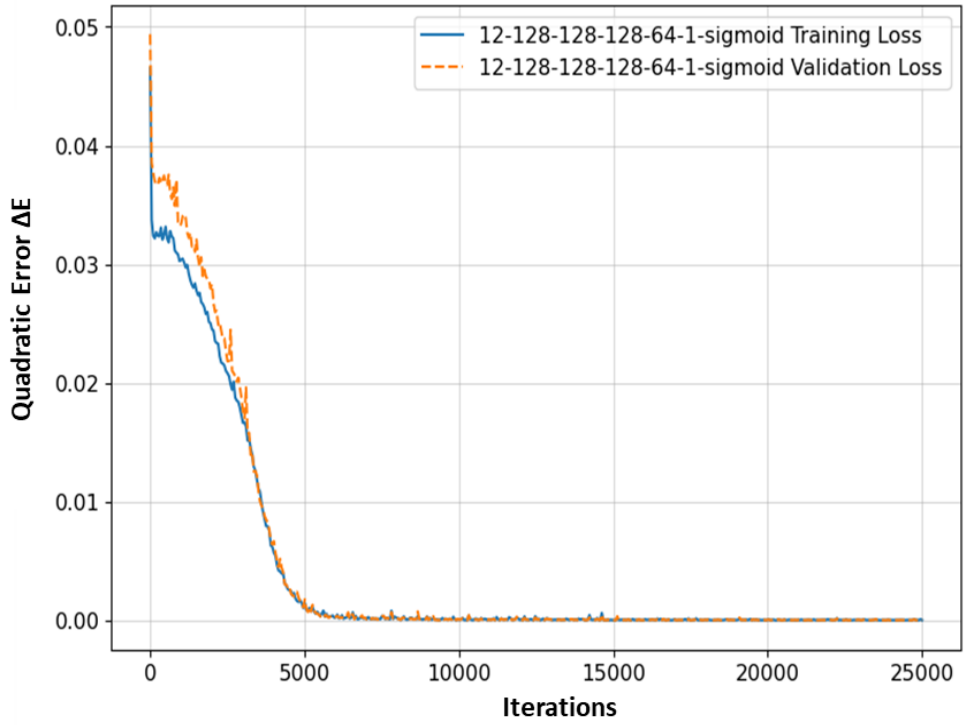

(c)

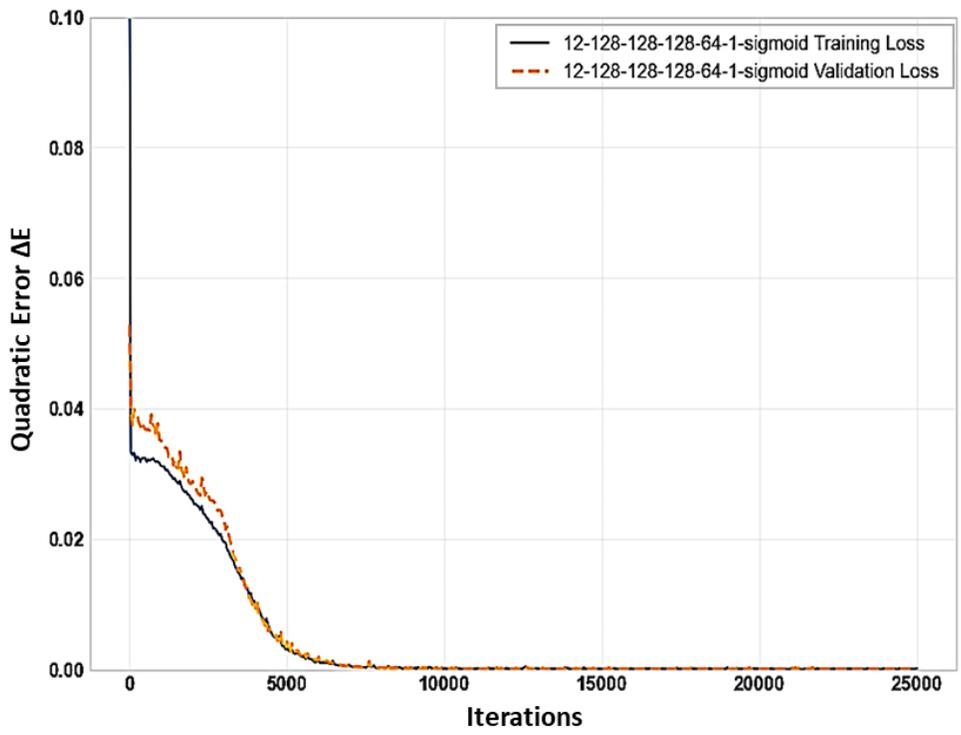

(d)

Figure 6. Learning curves for the ANN models during training and validation: (a) ANN (1) with layer configuration 16-8, (b) ANN (2) with layer configuration 144-256-64, (c) ANN (3) with layer configuration 128-128-128-64, and (d) ANN (3) evaluated with a data split of 80% for training and 20% for testing.

## 4.2. Comparison of the ANN models with other models

To evaluate the generalization performance and prediction accuracy of the models developed in this study, various metrics were used. Table 5 summarizes the results of the performance evaluation for all the tested models, including ANN models with varying hyperparameters, the RF model, and the SVM-RBF model.

Table 5. Metrics for the developed machine learning models

| Metrics | Model | | | | | |
|---|---|---|---|---|---|---|
| | SVM-RBF | RF | ANN (1) | ANN (2) | ANN (3) | ANN (3) optimized |
| MSE | 0.03 | $7.9034 \times 10^{-6}$ | 0.0148 | 0.0097 | 0.00003 | 0.000016 |
| RMSE | 0.17 | 0.0028 | 0.1216 | 0.0985 | 0.0055 | 0.004999 |
| r | 0.2828 | 0.3316 | 0.778 | 0.866 | 0.9996 | 0.9998 |
| $R^{2*}$ | 0.08 | 0.11 | 0.61 | 0.75 | 0.9993 | 0.9996 |

* $R^2$ value for the test set

Based on the *r* and *p*-values, six parameters with weak correlations were eliminated. The updated results for the modified RF model, which was evaluated using a dataset split into 80% for training (with 5-fold cross-validation) and 20% for testing, are as follows: the best $R^2$ score during cross-validation is 0.1236, while the $R^2$ scores on the validation and test sets are 0.12 and 0.11, respectively. The training score is notably higher at 0.56, indicating that despite the parameter adjustments, the model still exhibits a limited ability to generalize to unseen data, with potential overfitting observed.

For comparison, the RF model with 12 input parameters was also evaluated across different datasets, which were divided into three parts: 60% for training, 20% for validation, and 20% for testing. The R² value on the validation set was 0.10, while the R² value on the test set was 0.07. In contrast, the training score (R²) was 0.87. These metrics suggest that the model's generalization performance improved only slightly, with an R² value of 0.11 for the 6 input parameters compared to 0.07 for the 12 input parameters.

The MSE and RMSE for the SVM-RBF model are 0.03 and 0.17, respectively. The $R^2$ values for training, validation, and testing are 0.17, 0.09, and 0.08, respectively. The low $R^2$ values indicate that the SVM model with an RBF kernel does not capture the relationship between the input variables and the target variable well. This suggests that the model is not performing effectively. An $R^2$ value of 0.08 for the test set means that the model explains only 8% of the variance in the test data, confirming that the model does not generalize well to new data. These results suggest that neither the RF nor the SVM-RBF models effectively captured the correlation between the input variables and the target, indicating that these models may not be well-suited for this problem.

The ANN (1) model achieved an MSE of 0.0148, RMSE of 0.1216, $R^2$ of 0.61, and $r$ of 0.778. The ANN (2) model showed improvement with an MSE of 0.0097, RMSE of 0.0985, $R^2$ of 0.75, and $r$ of 0.866. The ANN (3) model, with its architecture modifications, demonstrated outstanding results: MSE of 0.00003, RMSE of 0.0055, $R^2$ of 0.9993, and $r$ of 0.9996. These results indicate that this architecture is highly effective for explaining the relationship between input variables of the EAF and the end-point phosphorous content of steel. Finally, the optimized ANN (3) model, which involved data set adjustments but retained the same architecture, achieved the following metrics: MSE of 0.000016, RMSE of 0.00499, $R^2$ of 0.9996, and $r$ of 0.9998.

To validate the ANN models, regression plots were used, as shown in Figure 7, to illustrate the relationship between the ANN model's outputs and the actual values. If the plotted data align closely with the dashed line, which represents the optimal linear fit between the actual and predicted values, it indicates that the training process was effective and the resulting errors are minimal. The closer the data points are to this dashed line, the smaller the errors between the actual and predicted values. A comparison of the scatterplots for the three ANN models shows that the model ANN (3)'s data distribution is significantly closer to the dashed line, with a high degree of overlap, compared to the other models.

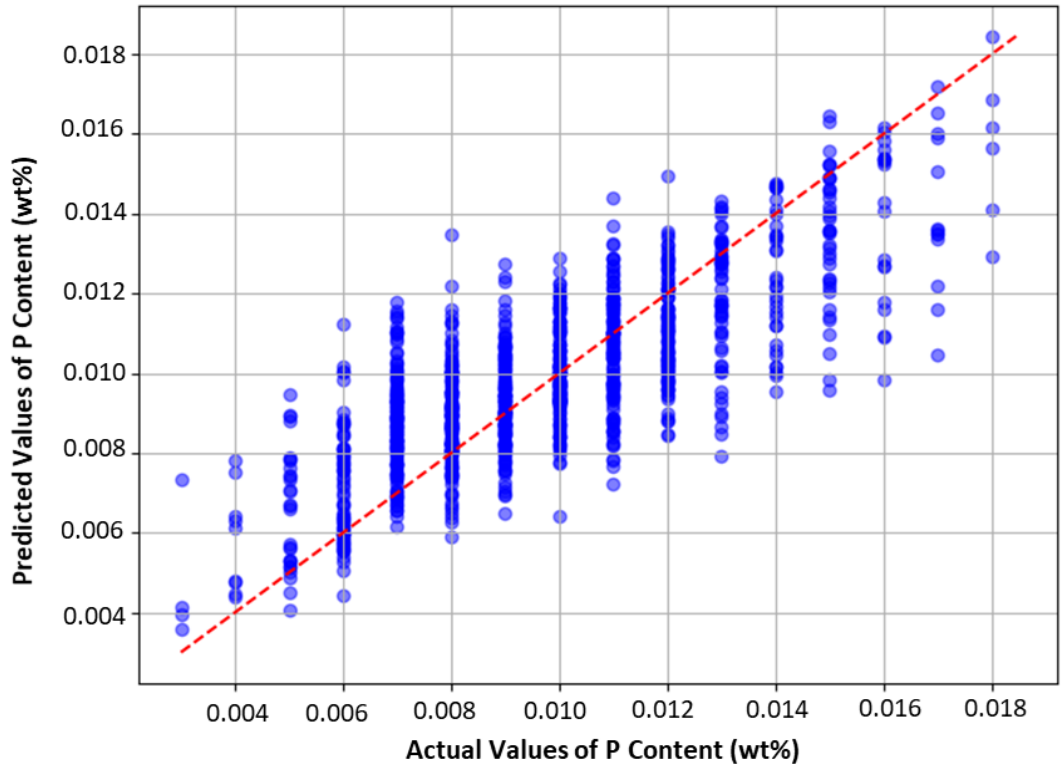

(a)

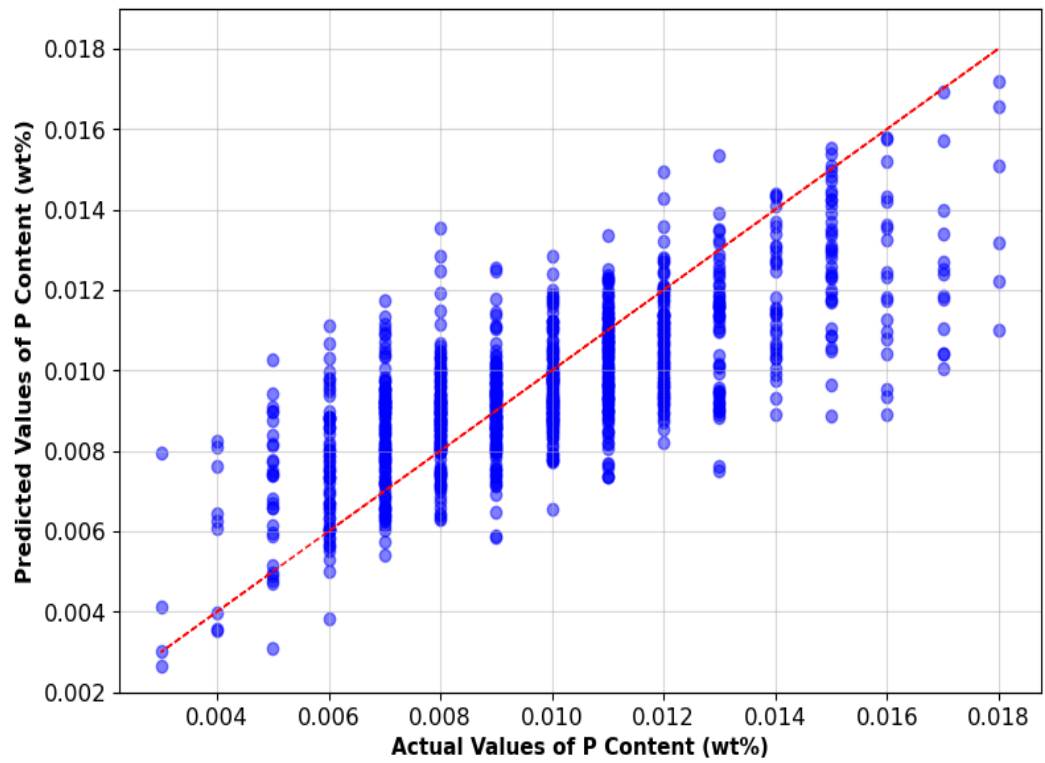

(b)

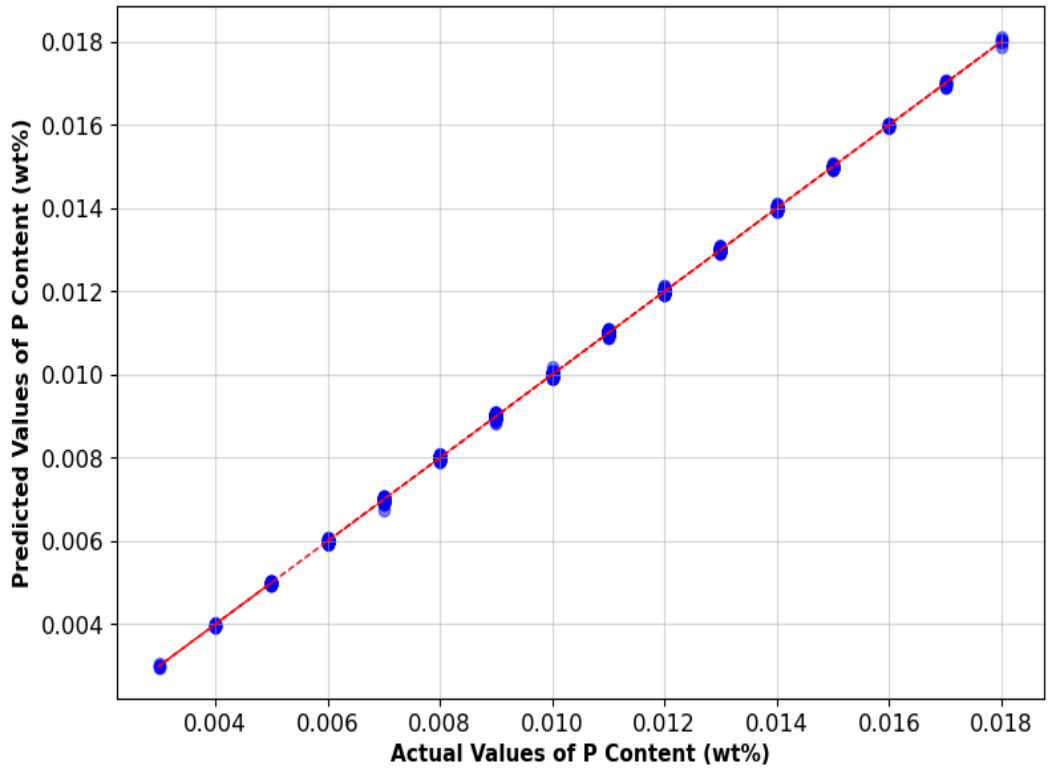

**(c)**

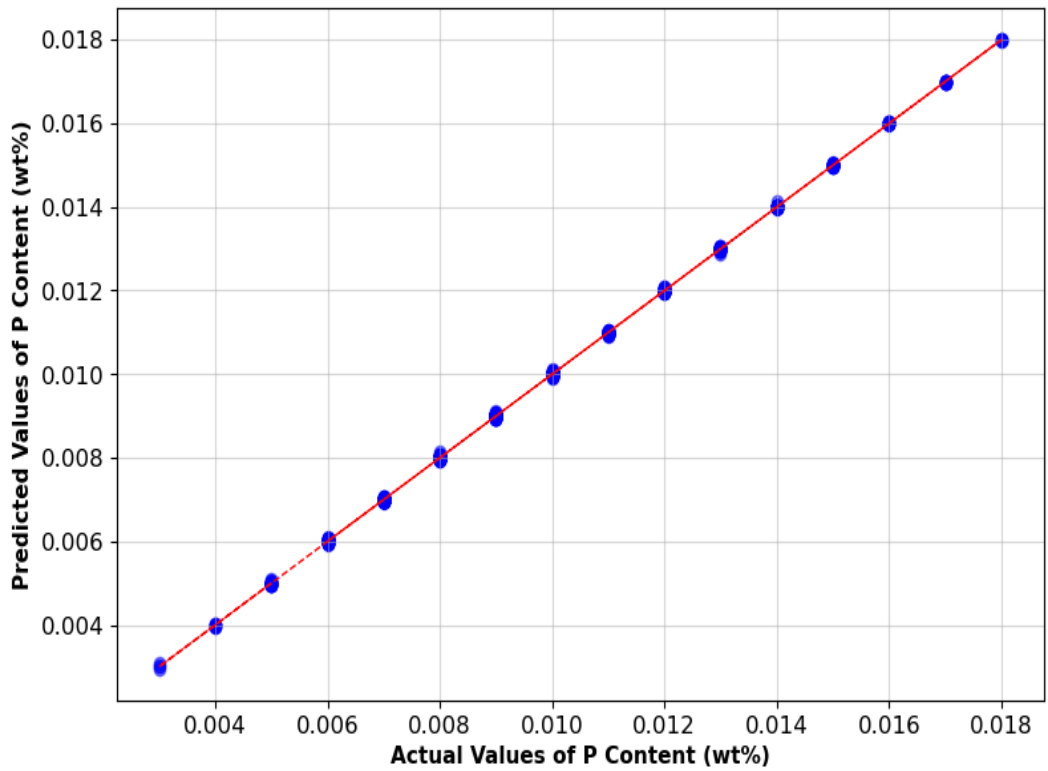

(d)

Figure 1. Comparison between predicted P values obtained by the ANN models and actual measured values: a) ANN (1): 16-8, b) ANN (2): 144-256-64, c) ANN (3): 128-128-128-64, and d) ANN (3) evaluated with a data split of 80% for training and 20% for testing.

Table 6 compares the performance metrics of the ANN models developed in this work with those of previous models in the literature, summarizing various studies that focus on different modeling approaches for predicting end-point phosphorus content in steel during EAF and BOF processes. It details the models employed, input parameters, dataset sizes, and evaluation metrics. The ANN (3) model, which used 12 input parameters, analyzed a dataset of 1,763 (with 1,005 utilized), achieving an $R^2$ of 0.9996 and an $r$ of 0.9998. In contrast, the ANN (2) model, which also applied to the same dataset, yielded an $R^2$ of 0.75 and $r$ of 0.866. Various models developed by Zhang et al. (2023) revealed moderate performance, with the highest $r$ being 0.608 for RF and the lowest at 0.382 for Ridge Regression. Both BPNN models by Zhou et al. (2022) showed $R^2$ values of 0.7596 and 0.8456, indicating decent predictive capabilities. The unhybrid ANN and hybrid physics-based ANN developed by Wang et al. (2022) showed NRMSE values of 0.1796 and 0.1775, respectively. Chang et al. (2021) developed various models and the $R^2$ values ranged from 0.280 for FCN to 0.729 for multi-channel GCN. He and Zhang [41] achieved an $r$ value of 0.79 for PCA-BPNN. Laha et al. [40] also developed various models for a reverberatory furnace, achieving an $R^2$ of 82% for the SVR model.

Table 6. Comparative summary of the newly proposed model and previous models

| References | Process | Model | Input parameters | Data size | Evaluation metrics |
|---|---|---|---|---|---|
| This Work | EAF (Scrap) | ANN (3) | 12 | 1763 (1005) | $R^2$: 0.9996<br>r: 0.9998 |
| | | ANN (2) | | | $R^2$: 0.75<br>r: 0.866 |
| Zou et al.[47] | EAF (HM + Scrap) | BPNN | 10 | 1250 (580) | -<br>- |
| Chen et al.[44] | EAF (HM + Scrap) | k means-BPNN-DT | 18 | 1258 (1114) | - |
| | | DNN | | | - |
| | | BPNN | | | - |
| Yuan et al.[34] | EAF | LS-SVM-PCR | 10 | 82 | - |
| Zhang et al.[46] | BOF | Ridge regression | 16 | 13000 (7776) | r: 0.382<br>MARE: 0.182<br>RMSE: 0.00369 |
| | | GBR | | | r: 0.599<br>MARE: 0.155<br>RMSE: 0.00325 |
| | | SVM | | | r: 0.52<br>MARE: 0.177<br>RMSE: 0.00342 |
| | | RF | | | r: 0.608<br>MARE: 0.156<br>RMSE: 0.00319 |
| | | CNN | | | r: 0.541<br>MARE: 0.173<br>RMSE: 0.00354 |
| Zhou et al.[51] | BOF | Unconstrained BPNN | 10 | (900) | $R^2$: 0.7596<br>RMSE: 0.0037 |
| | | Monotone-constrained BPNN | | | $R^2$: 0.8456<br>RMSE: 0.0030 |
| Wang et al.[48] | BOF | Unhybrid ANN | 19 | 28000 | NRMSE: 0.1796 |
| | | Hybrid physics-based ANN | | | NRMSE: 0.1775 |
| Chang et al.[43] | BOF | PLS | 42 | | $R^2$: 0.728<br>RMSE: 0.0019 |
| | | SVR | | | $R^2$: 0.622<br>RMSE: 0.0022 |
| | | FCN | | | $R^2$: 0.280 |

| | | ELM | | | RMSE: 0.0028 R$^2$: 0.620 |
| --- | --- | --- | --- | --- | --- |
| | | GCN | | | RMSE: 0.0022 R$^2$: -0.132 RMSE: 0.0038 |
| | | Multi-Channel GCN | | | R$^2$: 0.729 RMSE: 0.0019 |
| He and Zhang [41] | BOF | PCA and BPNN | 18 (7 with PCA) | 1978 | r: 0.79 |
| Laha et al. [40] | Reverberatory Furnace | RF, ANN, DENFIS, SVR | 10 | 54 | R$^2$: 82% (for SVR) |

HM stands for hot metal.

The hit rate measures the percentage of predictions within a specified error margin for the final phosphorus content of steel. Figure 8 illustrates the hit rates of the three ANN models developed in this work compared with those of models from the literature. For ANN (2), hit rates were 45% within ±0.001 wt% (10 ppm P), 72% within ±0.002 wt% (20 ppm P), 87% within ±0.003 wt% (30 ppm P), and 95% within ±0.004 wt% (40 ppm P). In contrast, both ANN (3) and the optimized ANN (3) with an 80%-20% split achieved a perfect hit rate of 100% across all error thresholds. Both ANN (2) and ANN (3) outperform earlier models [34, 44, 46, 47, 51], with ANN (3) showing particularly superior accuracy in predicting the final phosphorus content in steel.

ML models provide superior predictive accuracy compared to phenomenological models [32, 46]. They are generally easier to develop than mechanistic approaches like CFD, as they do not require a deep understanding of the underlying physical relationships, which are often based on simplified assumptions. However, a solid metallurgical understanding of the process is essential for the proper selection of features and input parameters. The dephosphorization process, in particular, presents challenges; as a physicochemical process occurring at high temperatures, it involves interconnected variables with complex, non-linear relationships. This complexity requires a model that can effectively capture these dynamics to achieve accurate predictions.

Comparing the present work to previously developed models, it is evident that neural network models can achieve high accuracy in predicting phosphorus content of steel. Among various neural network architectures—such as BPNN, CNN, GCN, ANN, and their derivatives—different configurations yield varying levels of predictive accuracy [41, 43, 44, 46-48, 50, 51]. For instance, our previous work showed that neither a DNN with 7 hidden layers and 416 neurons nor an ANN with 2 hidden layers and 24 neurons could match the accuracy of an ANN model with 4 hidden layers and 448 neurons [50]. This highlights the importance of carefully designing the architecture of neural network models.

However, the extensive hyperparameter optimization required, along with the need for substantial computing power and lengthy training times, can impose time constraints on their development and practical implementation. It is also necessary to develop a user-friendly interface that enables plant engineers and operators to utilize the model in situ. Moreover, the ML models have limited extrapolation capacity, and adapting them to different processes or plants often necessitates adjustments [32]. They do not enhance our understanding of the underlying physical phenomena. Thus, complementary physical models, such as thermodynamic and CFD models, are essential for gaining insights into process fundamentals.

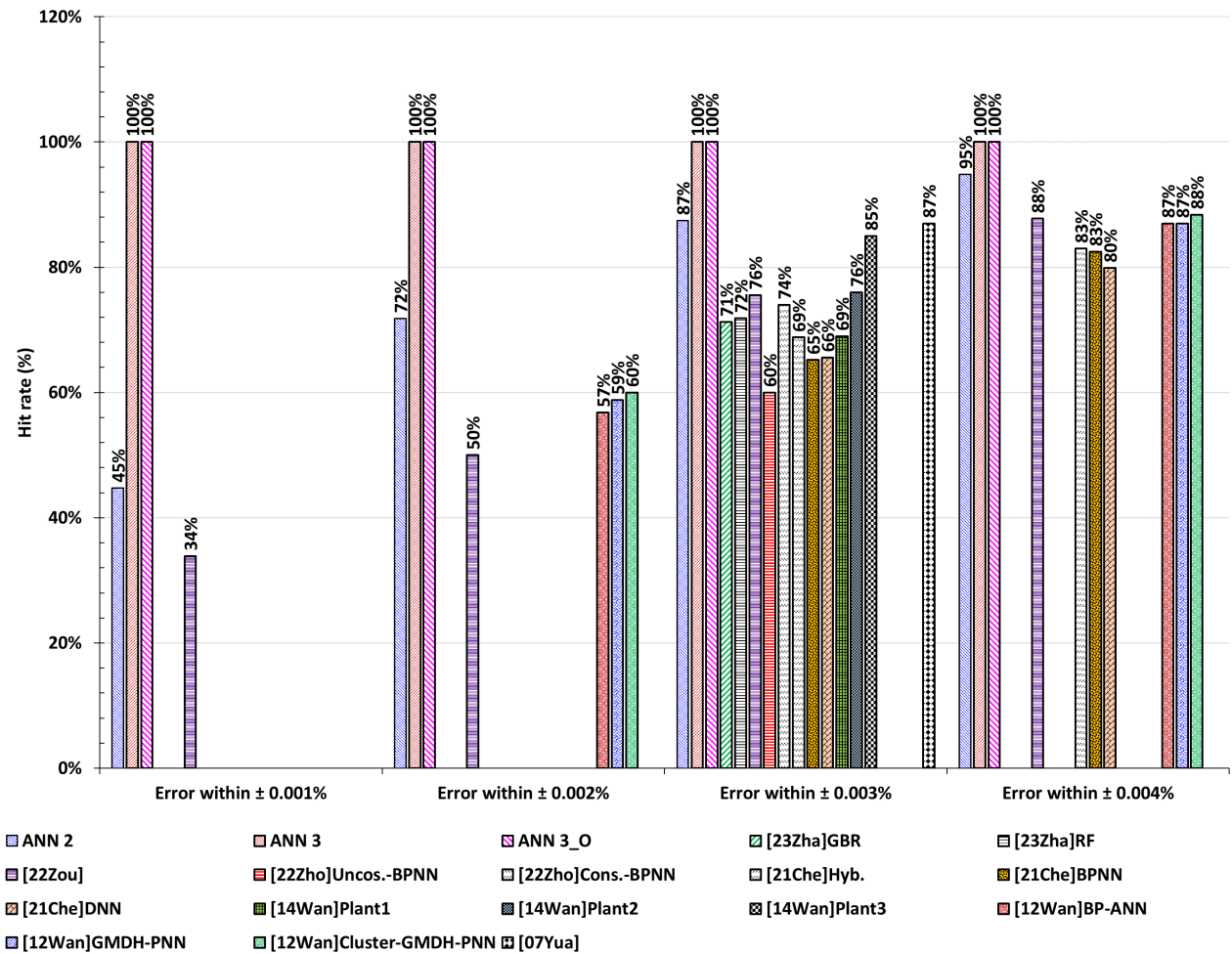

Figure 8. Hit rates of three ANN models developed in this work: ANN 1 (16-8), ANN 2 (144-256-64), ANN 3 (128-128-128-6), and ANN 3_O (80%-20% split) along with those of previously developed model [34, 44, 46, 47, 51, 66].

## 5. Conclusions

In the present work, various machine learning models were used to predict the phosphorus content in a low alloyed steel at the end of a scrap-based electric arc furnace (EAF) process. The tested models included a support vector machine (SVM) with a radial basis functions (RBF) kernel, a random forest (RF), and artificial neural networks (ANN). The main findings of this research are summarized below.

- Strong correlations with the end-point phosphorus content of steel were found for process duration, Cr and S contents in scrap, and injected oxygen ($p$-value < 0.01). Intermediate correlations were observed for scrap weight, Mn content in scrap, and injected lime (0.01 < $p$-value < 0.05). Weaker correlations were noted for energy consumption, deslagging temperature, C and Si contents in scrap, and tapping temperature ($p$-value > 0.05).
- Machine learning models, such as SVM-RBF and RF, did not yield satisfactory results in terms of phosphorus prediction accuracy. Several ANN models with different architectures were tested, and the most effective model consisted of 4 hidden layers and 448 neurons. This model was trained for 500 epochs with batches of 50 samples, and implemented using the TensorFlow library. Hyperparameters were carefully tuned to maximize performance, employing the Adam optimizer for adaptive learning rate adjustments and the sigmoid activation function to introduce non-linearity in each neuron.
- The optimized ANN model achieved remarkable performance compared to similar models reported in the literature, with a root mean square error (RMSE) of 0.004999, a mean squared error (MSE) of 0.000016, a correlation coefficient (*r*) of 0.9998, and a coefficient of determination ($R^2$) of 0.9996. Additionally, it demonstrated a perfect hit rate of 100% for predicting end-point phosphorus content within ±0.001 wt% in steel. These results confirm that, even with a limited amount of data (1005), an optimized ANN architecture can provide accurate and reliable predictions for the phosphorus content of steel in the EAF process.

**Acknowledgments:** The authors would like to acknowledge Finkl Steel-Sorel, Mitacs Accelerate Program (IT28458), and CIFAR for financial support of the project. In addition, the authors highly appreciate Finkl Steel-Sorel for providing the plant data and technical discussion throughout the project.

**Conflict of interest:** The authors declare that there are no conflicts of interest.

# References

1. Abadi, M.M., H. Tang, and M.M. Rashidi, *A Review of Simulation and Numerical Modeling of Electric Arc Furnace (EAF) and its Processes.* Heliyon, 2024.
2. Kildahl, H., et al., *Cost effective decarbonisation of blast furnace–basic oxygen furnace steel production through thermochemical sector coupling.* Journal of Cleaner Production, 2023. **389**: p. 135963.
3. WorldSteelAssociation. *Maximising scrap use helps reduce $CO_2$ emissions*. Raw materials 2024; Available from: https://worldsteel.org/steel-topics/raw-materials/.
4. Heo, J.H. and J.H. Park, *Effect of Slag Composition on Dephosphorization and Foamability in the Electric Arc Furnace Steelmaking Process: Improvement of Plant Operation.* Metallurgical and Materials Transactions B, 2021. **52**: p. 3613-3623.
5. Lin, W., et al., *A review of multi-phase slag refining for dephosphorization in the steelmaking process.* Frontiers in Materials, 2020. **7**: p. 602522.
6. Rodrigues, C., et al., *Effect of phosphorus content on the mechanical, microstructure and corrosion properties of supermartensitic stainless steel.* Materials Science and Engineering: A, 2016. **650**: p. 75-83.
7. Holappa, L. and A.C. Nava, *Secondary steelmaking*, in *Treatise on Process Metallurgy*. 2024, Elsevier. p. 267-301.
8. Menard, P. *Personal Communication*. 2023, Finkl Steel Sorel.
9. Compañero, R.J., A. Feldmann, and A. Tilliander, *Circular steel: how information and actor incentives impact the recyclability of scrap.* Journal of Sustainable Metallurgy, 2021. **7**: p. 1654-1670.
10. Anameric, B., D. Rohaus, and T.R. Riebeiro, *Ironmaking*, in *SME mineral processing & extractive metallurgy handbook*, R.C. Dunne, S.K. Kawatra, and C.A. Young, Editors. 2019: Society for Mining, Metallurgy, and Exploration (SME). p. 1781–1796.
11. Ripke, S.J., et al., *Iron ore beneficiation*, in *SME mineral processing & extractive metallurgy handbook*, R.C. Dunne, S.K. Kawatra, and C.A. Young, Editors. 2019: Society for Mining, Metallurgy, and Exploration (SME). p. 1755–1779.
12. Suito, H., R. Inoue, and M. Takada, *Phosphorus distribution between liquid iron and MgO saturated slags of the system $CaO-MgO-FeO_x-SiO_2$.* Tetsu-to-Hagané, 1981. **67**(16): p. 2645-2654.
13. Suito, H. and R. Inoue, *Effect of calcium fluoride on phosphorus distribution between MgO-saturated slags of the system $CaO-MgO-FeO_x-SiO_2$ and liquid iron.* Tetsu-to-Hagané, 1982. **68**(10): p. 1541-1550.
14. Suito, H. and R. Inoue, *Effects of $Na_2O$ and BaO additions on phosphorus distribution between $CaO-MgO-Fe_tO-SiO_2$-slags and liquid iron.* Transactions of the Iron and Steel Institute of Japan, 1984. **24**(1): p. 47-53.
15. Nakamura, S., F. Tsukihashi, and N. Sano, *Phosphorus partition between $CaO_{satd.}$-$BaO-SiO_2-Fe_tO$ slags and liquid iron at 1873 K.* ISIJ international, 1993. **33**(1): p. 53-58.
16. OI, O., et al., *Phosphate Capacity of the $CaO-CaF_2$ System Containing Chromium Oxide.* ISIJ International, 1994. **34**(10): p. 849-851.

17. Im, J., K. Morita, and N. Sano, *Phosphorus distribution ratios between CaO-SiO₂-FeₜO slags and carbon-saturated iron at 1573 K.* ISIJ international, 1996. **36**(5): p. 517-521.
18. Katsuki, J.-i., et al., *Removal of P and Cr by oxidation refining of Fe-36% Ni melt.* ISIJ international, 1996. **36**(Suppl): p. S73-S76.
19. Hamano, T. and F. Tsukihashi, *The Effect of B₂O₃ on Dephosphorization of Molten Steel by FeOₓ-CaO-MgOsatd.-SiO₂ Slags at 1873K.* ISIJ international, 2005. **45**(2): p. 159-165.
20. Lee, C. and R. Fruehan, *Phosphorus equilibrium between hot metal and slag.* Ironmaking & steelmaking, 2005. **32**(6): p. 503-508.
21. Li, G., T. Hamano, and F. Tsukihashi, *The effect of Na₂O and Al₂O₃ on dephosphorization of molten steel by high basicity MgO saturated CaO-FeOₓ-SiO₂ slag.* ISIJ international, 2005. **45**(1): p. 12-18.
22. Basu, S., A.K. Lahiri, and S. Seetharaman, *Phosphorus partition between liquid steel and CaO-SiO₂-P₂O₅-MgO slag containing low FeO.* Metallurgical and materials transactions B, 2007. **38**: p. 357-366.
23. Cho, M.K., J.H. Park, and D.J. Min, *Phosphate Capacity of CaO–SiO₂–MnO–FeO Slag Saturated with MgO.* ISIJ international, 2010. **50**(2): p. 324-326.
24. Li, F., et al., *Distribution ratios of phosphorus between CaO-FeO-SiO₂-Al₂O₃/Na₂O/TiO₂ slags and carbon-saturated iron.* Metallurgical and Materials Transactions B, 2017. **48**: p. 2367-2378.
25. Drain, P.B., et al., *Phosphorus partition and phosphate capacity of basic oxygen steelmaking slags.* ISIJ International, 2018. **58**(11): p. 1965-1971.
26. Heo, J.H. and J.H. Park, *Effect of direct reduced iron (DRI) on dephosphorization of molten steel by electric arc furnace slag.* Metallurgical and Materials Transactions B, 2018. **49**: p. 3381-3389.
27. Frueham, R.J., *AISI/DOE Technology Roadmap Program: Behavior of Phosphorus in DRI/HBI During Electric Furnace Steelmaking.* 2001, American Iron and Steel Institute (US).
28. Lee, M., D. Trotter, and O. Mazzei, *The production of low phosphorus and nitrogen steelsin an EAF using HBI.* Scandinavian journal of metallurgy, 2008. **30**(5): p. 286-291.
29. Hassan, A., et al. *Phosphorous behavior in Electric Arc Furnace steelmaking with the melting of high phosphorous content direct reduced iron.* in *METAL 2015-24th International Conference on Metallurgy and Materials, Conference Proceedings.* 2015.
30. Odenthal, H.J., et al., *Review on modeling and simulation of the electric arc furnace (EAF).* steel research international, 2018. **89**(1): p. 1700098.
31. Ek, M., et al., *New approach towards dynamic modelling of dephosphorisation in converter process.* Ironmaking & steelmaking, 2012. **39**(2): p. 77-84.
32. Hay, T., et al., *A review of mathematical process models for the electric arc furnace process.* steel research international, 2021. **92**(3): p. 2000395.
33. Tao, J. and W. Qian. *Intelligent method for BOF endpoint.* in *& [Mn] estimation,[in] 2006 6th World Congress on Intelligent Control and Automation, Dalian.* 2006.

34. Yuan, P., Z.-z. Mao, and F.-l. Wang, *Endpoint prediction of EAF based on multiple support vector machines.* Journal of Iron and Steel Research International, 2007. **14**(2): p. 20-24.
35. Zhaoyi, L., X. Zhi, and M. Hongji. *Prediction model of end-point phosphorous in converter based on cluster analysis and gray theory.* in *2008 7th World Congress on Intelligent Control and Automation*. 2008. IEEE.
36. Wang, H.-b., et al., *Prediction of endpoint phosphorus content of molten steel in BOF using weighted K-means and GMDH neural network.* Journal of Iron and Steel Research, International, 2012. **19**(1): p. 11-16.
37. Wang, R., et al., *Modeling Study of Metallurgical Slag Foaming via Dimensional Analysis.* Metallurgical and Materials Transactions B, 2021. **52**: p. 1805-1817.
38. Dong, Q., et al. *Research on relationship model of dephosphorization efficiency and slag basicity based on support vector machine.* in *2013 International Conference on Mechanical and Automation Engineering*. 2013. IEEE.
39. Liu, H., B. Wang, and X. Xiong, *Basic oxygen furnace steelmaking end-point prediction based on computer vision and general regression neural network.* Optik, 2014. **125**(18): p. 5241-5248.
40. Laha, D., Y. Ren, and P.N. Suganthan, *Modeling of steelmaking process with effective machine learning techniques.* Expert systems with applications, 2015. **42**(10): p. 4687-4696.
41. He, F. and L. Zhang, *Prediction model of end-point phosphorus content in BOF steelmaking process based on PCA and BP neural network.* Journal of Process Control, 2018. **66**: p. 51-58.
42. Elkoumy, M.M., et al., *Empirical Model for Predicting Process Parameters during Electric Arc Furnace Refining Stage Based on Real Measurements.* Steel Research International, 2019. **90**(11): p. 1900208.
43. Chang, S., et al., *Multi-channel graph convolutional network based end-point element composition prediction of converter steelmaking.* IFAC-PapersOnLine, 2021. **54**(3): p. 152-157.
44. Chen, C., N. Wang, and M. Chen, *Optimization of dephosphorization parameter in consteel electric arc furnace using rule set model.* Steel Research International, 2021. **92**(7): p. 2000719.
45. Klimas, M. and D. Grabowski, *Application of shallow neural networks in electric arc furnace modeling.* IEEE Transactions on Industry Applications, 2022. **58**(5): p. 6814-6823.
46. Zhang, R., et al., *Comparison of the Prediction of BOF End-Point Phosphorus Content Among Machine Learning Models and Metallurgical Mechanism Model.* Steel Research International, 2023. **94**(5): p. 2200682.
47. Zou, Y., et al., *Prediction Model of End-Point Phosphorus Content in EAF Steelmaking Based on BP Neural Network with Periodical Data Optimization.* Metals, 2022. **12**(9): p. 1519.
48. Wang, R., et al., *Hybrid method for endpoint prediction in a basic oxygen furnace.* Metals, 2022. **12**(5): p. 801.
49. Tomažič, S., et al., *Data-driven modelling and optimization of energy consumption in EAF.* Metals, 2022. **12**(5): p. 816.

50. Moosavi-Khoonsari, E., et al., *Controlling Minor Element Phosphorus in Green Electric Steelmaking Using Neural Networks*, in *REWAS 2025 at TMS 2025 Annual Meeting & Exhibition (Submitted)*. 2025.
51. Zhou, K.-X., et al., *Prediction model of end-point phosphorus content for BOF based on monotone-constrained BP neural network.* Journal of Iron and Steel Research International, 2022: p. 1-10.
52. Nenchev, B., et al., *Metallurgical data science for steel industry: A case study on basic oxygen furnace.* steel research international, 2022. **93**(12): p. 2100813.
53. Freuhan, J., *The making, shaping and treating of steel 11th edition—steelmaking and refining volume.* The AISE Steel Foundation, Pittsburgh, PA, 1998.
54. Maia, T.A. and V.C. Onofri, *Survey on the electric arc furnace and ladle furnace electric system.* Ironmaking & Steelmaking, 2022. **49**(10): p. 976-994.
55. Singh, R., *Applied welding engineering: processes, codes, and standards*. 2020, Butterworth-Heinemann. p. 33-38.
56. Rathaba, L.P., *Model fitting for electric arc furnace refining*. 2004: University of Pretoria (South Africa).
57. Busa, N., *Optimization of Steelmaking Processes in an Electric ARC Furnace*. 2023, Purdue University.
58. Kadkhodabeigi, M., H. Tveit, and S.T. Johansen, *Modelling the tapping process in submerged arc furnaces used in high silicon alloys production.* ISIJ international, 2011. **51**(2): p. 193-202.
59. Yang, X.-M., et al., *Critical evaluation of prediction models for phosphorus partition between CaO-based slags and iron-based melts during dephosphorization processes.* Metallurgical and Materials Transactions B, 2016. **47**: p. 2302-2329.
60. Nakamura, T., Y. Ueda, and T. Yanagase, *Optical Basicities in Some Oxide-Halide Systems.* ECS Proceedings Volumes, 1987. **1987**(1): p. 382.
61. Nassaralla, C. and R. Fruehan, *Phosphate capacity of $CaO-Al_2O_3$ slags containing $CaF_2$, $BaO$, $Li_2O$, or $Na_2O$.* Metallurgical Transactions B, 1992. **23**: p. 117-123.
62. Liu, Z., S.S. Cheng, and L. Wang. *Factors Influencing Dephosphorization of Low Carbon Steel in Converter*. in *Materials Science Forum*. 2021. Trans Tech Publ.
63. Oh, M.K. and J.H. Park, *Effect of fluorspar on the interfacial reaction between electric arc furnace slag and magnesia refractory: Competitive corrosion-protection mechanism of magnesiowüstite layer.* Ceramics International, 2021. **47**(14): p. 20387-20398.
64. Vieira, D., et al., *Slag evaluation to reduce energy consumption and EAF electrical instability.* Materials Research, 2016. **19**(5): p. 1127-1131.
65. Li, F., et al., *Phosphate Capacities of $CaO–FeO–SiO_2–Al_2O_3/Na_2O/TiO_2$ Slags.* High Temperature Materials and Processes, 2019. **38**(2019): p. 50-59.
66. Wang, Z., et al., *The Control and Prediction of End-Point Phosphorus Content during BOF Steelmaking Process.* steel research international, 2014. **85**(4): p. 599-606.
67. Probst, P., M.N. Wright, and A.L. Boulesteix, *Hyperparameters and tuning strategies for random forest.* Wiley Interdisciplinary Reviews: data mining and knowledge discovery, 2019. **9**(3): p. e1301.

68. Cervantes, J., et al., *A comprehensive survey on support vector machine classification: Applications, challenges and trends*. Neurocomputing, 2020. **408**: p. 189-215.
69. Roy, A. and S. Chakraborty, *Support vector machine in structural reliability analysis: A review*. Reliability Engineering & System Safety, 2023. **233**: p. 109126.
70. Learn, S.; Available from: https://scikitlearn.org/stable/auto_examples/svm/plot_rbf_parameters.html.
71. IBM. Available from: https://www.ibm.com/topics/neural-networks#:~:text=Every%20neural%20network%20consists%20of,own%20associated%20weight%20and%20threshold.
72. Raiaan, M.A.K., et al., *A systematic review of hyperparameter optimization techniques in Convolutional Neural Networks*. Decision Analytics Journal, 2024: p. 100470.
73. Bergstra, J. and Y. Bengio, *Random search for hyper-parameter optimization*. Journal of Machine Learning Research, 2012. **13**(2).
74. Karbowniczek, M., E. Kawecka-Cebula, and J. Reichel, *Investigations of the dephosphorization of liquid iron solution containing chromium and nickel*. Metallurgical and Materials Transactions B, 2012. **43**: p. 554-561.
75. Sigworth, G.K. and J.F. Elliott, *The thermodynamics of liquid dilute iron alloys*. Metal Science, 1974. **8**(1): p. 298-310.
76. Yang, D., et al., *Effect of $Cr_2O_3$ content on viscosity and phase structure of chromium-containing high-titanium blast furnace slag*. Journal of Materials Research and Technology, 2020. **9**(6): p. 14673-14681.
77. Ma, S., et al., *Effect of MnO content on slag structure and properties under different basicity conditions: A molecular dynamics study*. Journal of Molecular Liquids, 2021. **336**: p. 116304.
78. Kawa, Y. and H. Mayani, *Effect of alloying elements on the activity of phosphorous in molten iron*. Tetsu to Hagane, 1982. **vol 68**(11).
79. Dovoedo, Y. and S. Chakraborti, *Boxplot-based outlier detection for the location-scale family*. Communications in statistics-simulation and computation, 2015. **44**(6): p. 1492-1513.
80. Ratnasingam, S. and J. Muñoz-Lopez, *Distance correlation-based feature selection in random forest*. Entropy, 2023. **25**(9): p. 1250.
81. Dahiru, T., *P-value, a true test of statistical significance? A cautionary note*. Annals of Ibadan postgraduate medicine, 2008. **6**(1): p. 21-26.
82. Samarasinghe, S., *Neural networks for applied sciences and engineering: from fundamentals to complex pattern recognition*. 2016: Auerbach publications.